\documentclass[accepted]{uai2023} % after acceptance, for a revised
\usepackage[american]{babel}
\usepackage{natbib} % has a nice set of citation styles and commands
\usepackage{mathtools} % amsmath with fixes and additions
\usepackage{booktabs} % commands to create good-looking tables
\usepackage{tikz} % nice language for creating drawings and diagrams
\usepackage{authblk}

\usepackage[utf8]{inputenc} % allow utf-8 input
\usepackage[T1]{fontenc}    % use 8-bit T1 fonts
\usepackage{hyperref}       % hyperlinks
\usepackage{url}            % simple URL typesetting
\usepackage{booktabs}       % professional-quality tables
\usepackage{amsfonts}       % blackboard math symbols
\usepackage{nicefrac}       % compact symbols for 1/2, etc.
\usepackage{microtype}      % microtypography
\usepackage{xcolor}         % colors
\usepackage{colortbl}
\usepackage{amsmath}
\usepackage{multirow}
\usepackage{lipsum}
\usepackage{adjustbox}
\usepackage{placeins}
\usepackage{wrapfig}
\usepackage{subcaption}
\usepackage{algpseudocode}
\usepackage{algorithm}
\algnewcommand\algorithmicforeach{\textbf{for each}}
\algdef{S}[FOR]{ForEach}[1]{\algorithmicforeach\ #1\ \algorithmicdo}

\usepackage{amsmath}
\usepackage{amssymb}
\usepackage{amsthm}
{
\theoremstyle{plain}
      
}
\usepackage{eepic,epic}
\usepackage{epsfig}
\usepackage{graphicx}

\usepackage{scalerel,stackengine}
\stackMath
\newcommand\reallywidehat[1]{%
\savestack{\tmpbox}{\stretchto{%
  \scaleto{%
    \scalerel*[\widthof{\ensuremath{#1}}]{\kern.1pt\mathchar"0362\kern.1pt}%
    {\rule{0ex}{\textheight}}%WIDTH-LIMITED CIRCUMFLEX
  }{\textheight}% 
}{2.4ex}}%
\stackon[-6.9pt]{#1}{\tmpbox}%
}
\def\delequal{\mathrel{\ensurestackMath{\stackon[1pt]{=}{\scriptstyle\Delta}}}}

\newtheorem{thm}{Theorem}[section]

\newtheorem{lem}[thm]{Lemma}

\theoremstyle{definition}

\theoremstyle{remark}

\numberwithin{equation}{section}

\title{Phase-shifted Adversarial Training}
\author[1]{Yeachan Kim\href{mailto:<yeachan@deargem.me>?Subject=Your UAI 2023 paper}}
\author[2]{Seongyeon Kim\href{mailto:<synkim@kias.re.kr>?Subject=Your UAI 2023 paper}\thanks{Co-corresponding authors who equally contributed to this work.}}
\newcommand\CoAuthorMark{\footnotemark[\arabic{footnote}]}

\author[3]{Ihyeok Seo\href{mailto:<ihseo@skku.edu>?Subject=Your UAI 2023 paper}\protect\CoAuthorMark}
\author[4]{Bonggun Shin\href{mailto:<bonggun.shin@deargem.me>?Subject=Your UAI 2023 paper}\protect\CoAuthorMark}
\affil[1]{%
    Deargen Inc., Seoul, Republic of Korea
}
\affil[2]{%
    School of Mathematics, Korea Institute for Advanced Study, Seoul, Republic of Korea
}
\affil[3]{%
    Department of Mathematics, Sungkyunkwan University, Suwon, Republic of Korea
}
\affil[4]{%
    Deargen USA Inc., Atlanta, GA
}
\begin{document}
\maketitle
\begin{abstract}
Adversarial training (AT) has been considered an imperative component for safely deploying neural network-based applications. However, it typically comes with slow convergence and worse performance on clean samples (i.e., non-adversarial samples). In this work, we analyze the behavior of neural networks during learning with adversarial samples through the lens of response frequency. Interestingly, we observe that AT causes neural networks to converge slowly to high-frequency information, resulting in highly oscillatory predictions near each data point. To learn high-frequency content efficiently, we first prove that a universal phenomenon, the frequency principle (i.e., lower frequencies are learned first), still holds in AT. Building upon this theoretical foundation, we present a novel approach to AT, which we call phase-shifted adversarial training (PhaseAT). In PhaseAT, the high-frequency components, which are a contributing factor to slow convergence, are adaptively shifted into the low-frequency range where faster convergence occurs. For evaluation, we conduct extensive experiments on CIFAR-10 and ImageNet, using an adaptive attack that is carefully designed for reliable evaluation. Comprehensive results show that PhaseAT substantially improves convergence for high-frequency information, thereby leading to improved adversarial robustness.

\end{abstract}

\section{Introduction}\label{sect:intro}

\begin{figure}[t]
\centering
\subfloat[Low-frequency errors]{%
\centering
  \includegraphics[width=0.99\linewidth]{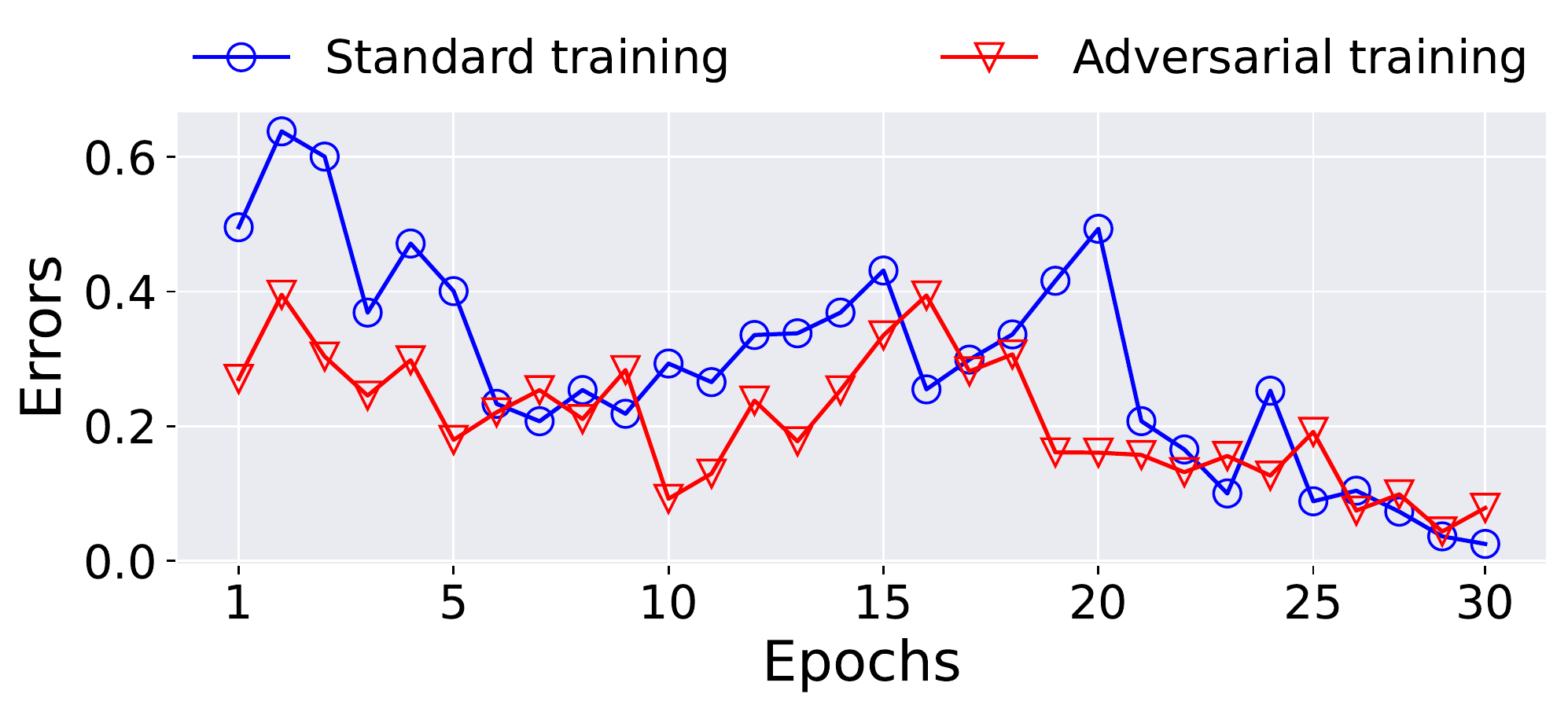}%
}
% \vspace{0.3cm}

\subfloat[High-frequency errors]{%
\centering
  \includegraphics[width=0.99\linewidth]{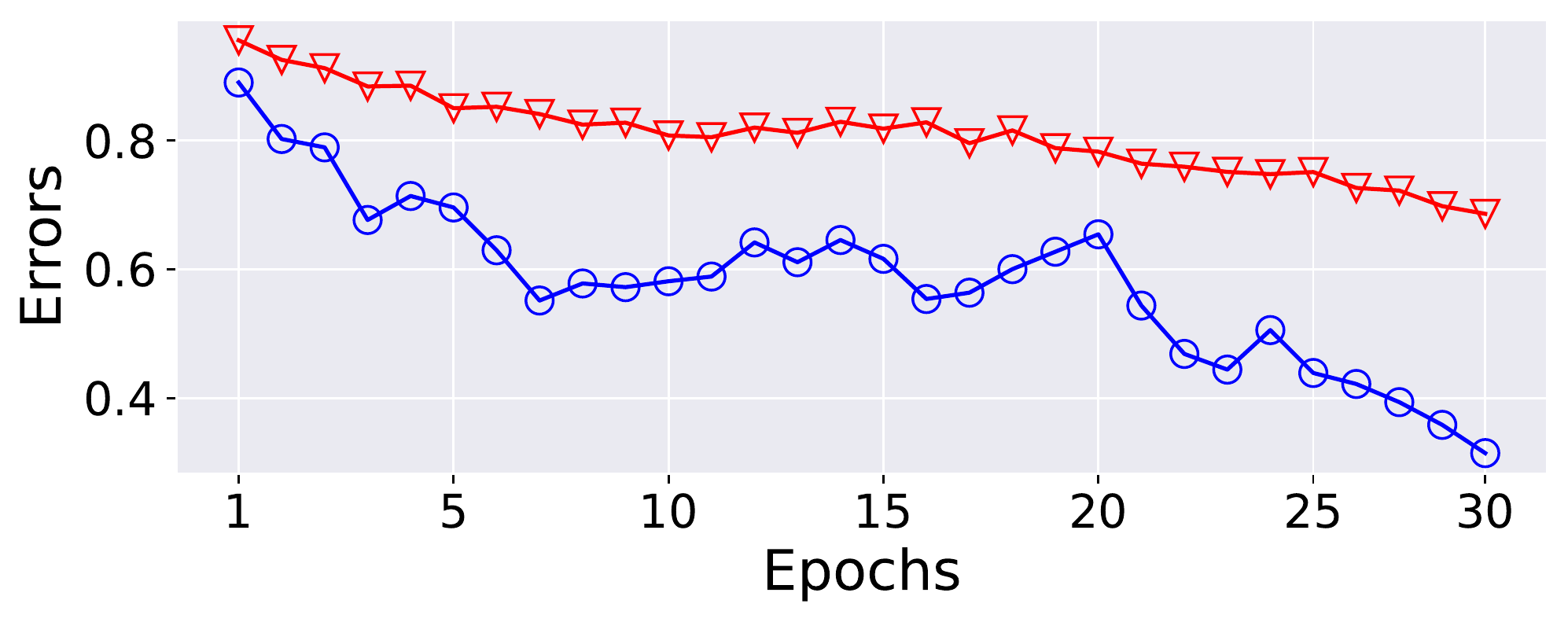}
}
\caption{Errors of frequency components (high, low) between the training dataset and neural networks. Here, we use CIFAR-10 dataset for validation.}
% This indicates that adversarial training causes neural networks to have the low convergence on high-frequency information.
\label{fig:motivation}
\end{figure}

Despite the remarkable success, deep neural networks are known to be susceptible to crafted imperceptible noise called \textit{adversarial attacks} \citep{szegedy2013intriguing}, which can have severe consequences when deployed in critical applications such as self-driving cars, medical diagnosis, and surveillance systems. In response to such negative implications, there has been a recent surge in research aimed at preventing adversarial attacks, such as adversarial training \citep{madry2017towards,wong2019fast,sriramanan2020guided,sriramanan2021towards}, data augmentation \citep{gong2021maxup,wang2021augmax,rebuffi2021fixing}, and regularization \citep{qin2019adversarial}. Adversarial training is considered one of the most effective ways to achieve adversarial robustness. The early attempt generates the attack by only a single gradient descent on the given input, which is known as fast gradient sign method (FGSM) \citep{goodfellow2014explaining}. However, it was later shown to be ineffective against strong multiple-step attacks \citep{kurakin2016adversarial}, e.g., projected gradient descent (PGD). Hence a myriad of defense strategies are introduced to build the robust models against the strong attacks by injecting the regularization to the perturbations \citep{sriramanan2020guided,sriramanan2021towards}, randomly initializing the perturbations \citep{wong2019fast,tramer2019adversarial}, and increasing the number of update steps for the perturbation to approximate the strong attack \citep{madry2017towards,zhang2019theoretically}.

Instead of introducing new attack and defense strategies, we analyze adversarial training through the lens of the frequency of the general mapping between inputs and outputs (e.g., neural networks). For this purpose, we calculate the errors of the frequency components between the dataset and neural networks to observe how the networks converge to the labeling function of the training dataset in terms of the frequencies. Compared to standard training (blue dots in Figure \ref{fig:motivation}), we find that adding adversarial examples (red inverted triangles in Figure \ref{fig:motivation}) causes the model to slowly converge in the high-frequency components (Figure \ref{fig:motivation}(b)). \footnote{We use the filtering method \citep{xu2019frequency} which explicitly splits the frequency spectrum into the high and low components and calculates the frequencies by applying the Fourier transform of the Gaussian function (please refer to Supplementary material for more detailed information).} 
This indicates that adversarial robustness comes at significantly increased training time compared to standard training.

To learn high-frequency contents of the dataset efficiently, phase-shift deep neural networks (PhaseDNN) \citep{cai2020phase} is proposed based on a universal phenomenon of frequency principle (F-principle) \citep{xu2019training,xu2019frequency,rahaman2019spectral,luo2019theory}, that is, \textit{deep neural networks often fit target functions from low to high frequencies during the training}. PhaseDNN learns high-frequency contents by shifting these frequencies to the range of low-frequency to exploit the faster convergence in the low-frequencies of the DNN. However, it is challenging to apply PhaseDNN to adversarial training because PhaseDNN was optimized to solve a single-dimensional data (e.g., electromagnetic wave propagation, seismic waves).

In this work, we propose phase-shifted adversarial training (PhaseAT) to achieve adversarial robustness efficiently and effectively. To this end, we theoretically prove that the F-principle holds not only in standard training but also in adversarial training. We then extend the phaseDNN to adopt high-dimensional data and learn the adversarial data effectively. In summary, our contributions include the following:
\begin{itemize}
    \item We conduct a pioneering analysis of adversarial training from the perspective of response frequency.
    
    \item We establish a mathematical foundation for how adversarial training behaves by considerably extending a universal phenomenon of frequency principle.  
    
    \item We propose a new phase-shifted adversarial training algorithm based on our proposed theory, outperforming other strong baselines in many different settings.
\end{itemize}
% For evaluations, we design adaptive attacks to reliably estimate the robustness of the proposed method, and comprehensive results clearly show that the faster and more accurate learning of high-frequency components leads to the strongest adversarial robustness compared to both iterative and non-iterative defense strategies. Our contributions include the following:

% Theoretic background (proving)

\section{Background of Adversarial Training}
% In this paper, we consider the adversarial robustness against $\ell_{\infty}$ constraiend adversaries. Hence the 
Adversarial training is a method for learning networks that are robust to adversarial attacks. Given a network $\mathcal{T}$ parameterized by $\theta$, a dataset $\{x_j, y_j\}_{j=0}^{N-1}$ where $N$ is the size of dataset, a loss function $\ell$ and a threat model $\Delta$, the learning problem can be cast to the following robust optimization problem.
\begin{equation}\label{eq:adv_train}
    \min_{\theta}\sum_{j=0}^{N-1}\max_{\delta \in \Delta}\ell(\mathcal{T}(x_j + \delta), y_j)
\end{equation}
where $\delta$ is the adversarial perturbation. As we consider the adversarial robustness against $L_{\infty}$ constrained adversaries, the threat model $\Delta$ takes the perturbation $\delta$ such that $\Vert \delta \Vert_{\infty} \le \epsilon$ for some $\epsilon > 0$. For adversarial training, it is common to use adversarial attack to approximate the inner maximization over $\Delta$, followed by some variation of gradient descent on the model parameters $\theta$. For example, FGSM attack \citep{goodfellow2014explaining} has the following approximation.
\begin{equation}\label{eq:adv_train}
    \delta = \epsilon \cdot \text{sign}(\nabla_{x}\ell(\mathcal{T}(x_j), y_j))
\end{equation}

% \subsection{F-principle}
% % frequency definition
% \lipsum[1]

\section{F-principle in Adversarial Training}\label{sect:fprinciple}
This section is devoted to theoretically demonstrating the F-principle in adversarial training as well as standard training (i.e., $\delta=0$). 
We first represent the total loss $L(\theta)$ in the frequency domain and then quantify the rate of change of $L(\theta)$ contributed by high frequencies. 
We provide detailed proof in Supplementary material.

\subsection{The total loss in the frequency domain}
Given a dataset $\{x_j, y_j\}_{j=0}^{N-1}$, $\mathcal T_\theta$ is the DNN output\footnote{Here we write $\mathcal T_\theta$ instead of $\mathcal T$ to explicitly denote the dependency on parameter $\theta$.} and $g(x)$ is the target function (also known as labeling function) such that $g(x_j)=y_j$. 
Then the total loss $L$ is generally defined by
$$L(\theta)=\frac{1}{N}\sum_{j=0}^{N-1} \ell (\mathcal T_{\theta},g)(x_j).$$
%where $\ell \big(\mathcal T_{\theta}, g\big)(x)$ is a loss function at each data point $x\in\mathbb{R}^d$.
Here, for example,
$\ell (\mathcal T_{\theta}, g )(x) = \|\mathcal T_{\theta}(x)-g(x)\|^2$
for mean-squared error loss function, and for cross-entropy loss function
$\ell (\mathcal T_{\theta}, g)(x)=-g(x) \cdot \log \mathcal T_{\theta}(x)$
where the log function acts on each vector component of $\mathcal T_\theta$.
In adversarial training, we define an \textit{adversarial function} $\mathcal A:\mathbb{R}^d \rightarrow \mathbb{R}^d$ by $\mathcal{A}(x)=x+\delta(x)$ with the adversarial perturbation $\delta$ and the corresponding output is given by $\mathcal T_\theta\circ\mathcal A$.
In reality, it may be considered that $\mathcal T_\theta$ and $g$ are bounded in a compact domain containing $\{x_j\}_{j=0}^{N-1}$. Then for the two common examples, $\ell (\mathcal{T}_\theta\circ\mathcal A, g)$ is absolutely integrable and $\ell (\mathcal{T}_\theta\circ\mathcal A,g)$ is differentiable with respect to the first argument. 
In this regard, these properties are considered to be
possessed by a loss function generally.

\begin{thm}\label{thm2}
Let $\widehat{f}$ denote the Fourier transform of $f$. 
Then, in the frequency domain
\begin{equation}\label{loss'} 
L(\theta)=\lim_{\varepsilon \rightarrow 0} \frac{1}{N} \sum_{j=0}^{N-1}\int_{\mathbb{R}^d} e^{2\pi i x_j \cdot\xi } e^{-\pi  \|\varepsilon\xi\|^2} \reallywidehat{\ell(\mathcal T_{\theta}\circ \mathcal{A},g)}(\xi) d\xi.
\end{equation}
\end{thm}

This representation is inspired by a convergence property, $G_\varepsilon \ast L(\theta)\rightarrow L(\theta)$, of convolution with approximate identities $G_{\varepsilon}(x):=\varepsilon^{-d}e^{-\pi\|\varepsilon^{-1}x\|^2}$. See Lemma D.1 in Supplementary material.

Now we split the integral in Eq.~\ref{loss'} into the region $\|\xi\|\leq \eta$ and $\|\xi\|\geq \eta$ for any $\eta>0$ to separate the total loss into two parts $L_{\leq\eta}(\theta)$ and $L_{\geq\eta}(\theta)$ contributed by low and high frequencies, respectively:
\begin{equation}\label{XX}
 L(\theta)
= \lim_{\varepsilon \rightarrow 0} \big( L_{\leq\eta}(\theta) +  L_{\geq\eta}(\theta)\big).
\end{equation}

\subsection{Quantifying the rate of change in total loss}
Let $W^{s,\infty}(\mathbb{R}^d)$ denote the Sobolev spaces\footnote{For a vector-valued $f$, we also write $f\in W^{s,\infty}(\mathbb{R}^d)$ to represent all of its component functions are contained in the space.} 
 which consist of bounded functions having bounded derivatives up to a given order $s$, and thus the index $s$ reflects the degree of regularity. 
\begin{thm}\label{thm}
Consider a DNN with multi-layer in adversarial training. If an activation function $\sigma$ and the target function $g$ satisfy $\sigma \in W^{s,\infty}(\mathbb{R})$ and $g \in W^{2s,\infty}(\mathbb{R}^d)$ for some $s\in \mathbb{N}$, then there exists a constant $C>0$ independent of $s,\eta$ such that for any $\eta>0$
\begin{align}\label{Fp}
\nonumber
\|\nabla_{\theta}&L(\theta)-\nabla_{\theta}L_{\leq\eta}(\theta)\|
\\&\qquad\,\,\approx\|\nabla_{\theta}L_{\geq\eta}(\theta)\| \leq C \max\{N,d^d\} \eta^{-2s}.
\end{align}
\end{thm}

Theorem \ref{thm} implies that the rate of decrease in the corresponding high-frequency region loss ($L_{\geq\eta}(\theta)$) is much greater than the rate of increase in $\eta$.
In other words, a model tends to fit target functions from low to high frequency information during the adversarial training. 
The mathematical insight of this theorem is that the regularity of a network converts into the decay of the total loss in the frequency domain.
Specifically, a network with a more regular activation function (bigger $s$) converges more slowly, according to $\eta^{-2s}$ in Eq.~\ref{Fp}. As the frequency increases, this becomes more evident.
For example, if $\sigma$ is ReLU or eLu then $s=1$ or $s=2$, respectively. For $\tanh$ and sigmoid activation functions, $s$ may be taken arbitrarily large.

% \textcolor{red}{The approximation in Eq.~\ref{Fp} becomes more accurate as $\varepsilon$ diminishes smaller in Eq.~\ref{XX}, and
% $\varepsilon=\min\{1/\sqrt[d]{N}, 1/ d\}$ is taken inversely proportional to the size $N$ of dataset or the dimension $d$ of input data. 
% In this way, when $N$ or $d$ is large, $\varepsilon$ becomes small.
% We motivate on two facts concerning our choice of $\varepsilon$; real-life datasets are generally high-dimensional (e.g., images and languages) and neural networks learn well when the size of dataset is large.}

The approximation in Eq.~\ref{Fp} becomes more accurate as $\varepsilon$ diminishes in Eq.~\ref{XX}, and
$\varepsilon=\min\{1/\sqrt[d]{N}, 1/ d\}$ is inversely proportional to the size of dataset ($N$) or the dimension of input data ($d$). 
Therefore, when $N$ or $d$ are large, $\varepsilon$ decreases. This is a common phenomenon in real-world datasets, which are typically high-dimensional and contain a large number of data samples (e.g., images and languages).

% \textcolor{red}{Our novel theory is distinct from its ancestors \citep{xu2019frequency,luo2019theory} in two folds. 
% Firstly, we further generalize F-principle by showing that it holds for the cross-entropy loss. 
% Secondly, we show the principle prominently by giving a faster decay rate, with $2s$ in Eq.~\ref{Fp}, than previously known. 
% Lastly, we provide the mathematical justification that F-principle holds in adversarial training settings.
% This is an important contribution because the high-frequency information is richer in adversarial training than in general deep learning settings.}

Our innovative theory differs from its predecessors \citep{xu2019frequency,luo2019theory} in two ways.
First, we generalize F-principle by showing that it holds for the cross-entropy loss.
Second, we provide a faster decay rate with $2s$ in Eq.~\ref{Fp}, which serves as one of the motivations for using phase-shifting, particularly in adversarial training.
Finally, we provide the mathematical justification for the F-principle in adversarial training settings.

\section{Phase-shifted Adversarial Training}

In this section, we detail the proposed method, coined phase-shifted adversarial training (PhaseAT). We first present the existing PhaseDNN \citep{cai2020phase} and its limitations (Section~\ref{subsect:phasednn}). We then redesign the original PhaseDNN to make it more practical and suitable for adversarial training (Section~\ref{subsect:mphasednn}). Finally, we elaborate PhaseAT by optimizing PhaseDNN through adversarial training (Section~\ref{subsect:advphasednn}).

\subsection{Phase Shift Deep Neural Networks}\label{subsect:phasednn}

To learn highly oscillatory functions efficiently, \citep{cai2020phase} propose PhaseDNN which shifts the high-frequency components of the dataset to the range of the low-frequency range for fast convergence because according to F-principle, neural networks learn low-frequency information first before learning high-frequency components. On the phase-shifted dataset, the neural networks learn the target function with low-frequency contents. Unfortunately, this requires that the frequency extraction of the original training data has to be done numerically using convolutions with a frequency selection kernel, which requires a huge memory footprint, i.e., a storage of $O$(N × N). Alternatively, the neural networks can be learned on the data from all ranges of frequencies while the phase-shift is included in the makeup of the PhaseDNN. This version of PhaseDNN is called \textit{coupled PhaseDNN}, and we adopt this version to avoid the cost of decomposing the frequency components on the original dataset.

% separate networks 
Since the phase-shift is performed on the networks rather than the dataset, PhaseDNN consists of an ensemble of networks, each of which is dedicated to a specific frequency. To learn higher frequency components, phase-shift is applied to each network separately. The output of PhaseDNN can be represented as follows:
\begin{equation}\label{eq:phasednn_first}
    \mathcal{T}(x) = \sum_{m=0}^{M-1}e^{i\omega_m x}\cdot \mathcal{T}_m(x)
\end{equation}
where $M$ is the size of ensemble, $T_{m}(x)$ represents one of the networks in the ensemble, and $\omega_m$ is the specific frequency for the networks $T_m$. Let the labeling function of the dataset be $g(\cdot)$, the phaseDNN is optimized by following the least square errors:
\begin{equation}
    \sum_{j=0}^{N-1}|g(x_j)-\mathcal{T}(x_j)|^2 = \sum_{j=0}^{N-1}\left\vert g(x_j)-\sum_{m=0}^{M-1}e^{i\omega_m x_j} \cdot \mathcal{T}_m(x_j)\right\vert^2
\end{equation}
Note that we always include the lowest frequency in the selected frequency, i.e., $\omega_0 = 0$, because the low frequencies typically dominates in real datasets \citep{xu2019frequency}.

\subsection{Multi-headed PhaseDNN}\label{subsect:mphasednn}

Applying the previous PhaseDNN to adversarial training has a few challenges. First, PhaseDNN was designed to learn single-dimensional data; the number of all possible frequencies grows exponentially with the dimension of the input, which prevents the use of high-dimensional inputs (e.g., images and languages). Second, PhaseDNN requires multiple networks to perform the phase-shift of the different frequencies, causing a large memory footprint.

To address the first challenge, we project the given inputs to the first principal component and use the projected scalar values to represent the original data. The forward propagation of PhaseDNN (Eq. \ref{eq:phasednn_first}) is reformulated as:
\begin{equation}\label{eq:projection}
    \mathcal{T}(x_j) = \sum_{m=0}^{M-1}e^{i\omega_m (x_j \cdot p)} \cdot \mathcal{T}_m(x_j)
\end{equation}
where $p$ is the first principal component of the input space calculated on the training dataset. Instead of observing all frequencies of high-dimensional data, we focus on the frequencies of the data along the first principal component $p$. Additionally, we bound the product result by normalizing the two vectors (i.e., $x$ and $p$) and multiplying the constant $C$ to fix the range of data points.

Lastly, we introduce a multi-headed PhaseDNN to avoid using the ensemble, which consumes large computational resources. We make each network $\mathcal{T}_m$ share the feature extracting networks and has frequency-dedicated networks for predictions. Thus, Eq~\ref{eq:projection} is reformulated as follows:
\begin{equation}\label{eq:phase_head}
    \mathcal{T}(x_j) = \sum_{m=0}^{M-1}e^{i\omega_m (x_j \cdot p)} \cdot \mathcal{H}_m(\mathcal{F}(x_j))
\end{equation}
where $\mathcal{F}(\cdot)$, $\mathcal{H}_m(\cdot)$ are the shared networks and the $m$-th frequency-dedicated classifier, respectively. It is worth noting that the classifiers $\mathcal{H}$ only make up a small portion of the total parameters, allowing PhaseDNN to efficiently learn highly oscillatory functions with a lower memory footprint.

\subsection{PhaseDNN with Adversarial Training}\label{subsect:advphasednn}
% \textcolor{red}{PhaseDNN requires dedicated frequencies $\{\omega_m\}_{m=0}^{M-1}$ for each head $\{\mathcal{H}_m\}_{m=0}^{M-1}$ which are shifted during the training.} 
% As networks should not reveal a large difference between clean data and adversarial data of the same data, we select the frequencies having the largest discrepancy of Fourier coefficients between the clean dataset, i.e., $\{x_j\}_{j=0}^{N-1}$ and the adversarial version of the dataset, i.e., $\{x_j = x_j + \delta\}_{j=0}^{N-1}$. 
% However, 
PhaseDNN requires shift frequencies ($\{\omega_m\}_{m=0}^{M-1}$) for each head ($\{\mathcal{H}_m\}_{m=0}^{M-1}$) used during training. The following explains how to choose those frequencies.
As the goal of training a network is to minimize the differences between a clean ($\{x_j\}_{j=0}^{N-1}$) and an adversarial version ($\{x_j = x_j + \delta\}_{j=0}^{N-1}$) of the same data point, we choose the target frequencies with the largest difference in Fourier coefficients between the two. In practice, estimating the Fourier coefficients of the total dataset in every optimization step requires huge computational resources; therefore, we approximate this by estimating the exponential moving average of the coefficients on mini batches. The discrepancy of the frequency $k$ between clean and adversarial batches is determined as follows:
\begin{equation}\label{eq:freq_select1}
    {d}_k = \left\vert \mathcal{F}_{k}(X) - \mathcal{F}_{k}(X+\Delta)\right\vert
\end{equation}
where $X$ and $\Delta$ indicate the batched data and its corresponding perturbations, respectively, $\mathcal{F}_k(\cdot)$ is the Fourier coefficient which is obtained as follows:
\begin{equation}\label{eq:freq_select2}
    \mathcal{F}_{k}(X) = \sum_{j=0}^{B-1}\mathcal{T}(X_j) \cdot e^{-2 \pi i k (X_j \cdot p)}
\end{equation}
where $B$ is the batch size. The estimated discrepancy $d_k$ for all frequencies is then used to derive the multinomial distribution to sample the frequencies $\omega_m$ for each head of the phaseDNN\footnote{Similar to the previous work, we always include the zero frequency for the one of the head networks.}. The reason for sampling frequencies from a multinomial distribution is that the training dataset is constantly changing during adversarial training. In this case, a fixed set of frequencies (e.g., peaks of frequencies used in \citep{xu2019frequency}) does not accurately reflect the critical frequency information. As a result, by stochastically learning the frequency differences, the model could decrease the prediction differences between clean and adversarial data.

\paragraph{Attack generation.} With the practically-designed PhaseDNN, we perform adversarial training, which causes the neural networks to highly oscillate near the training data. Motivated by the recent finding that FGSM-based training can sufficiently prevent strong attacks \citep{wong2019fast}, we train PhaseDNN to be robust against FGSM attacks.
% \footnote{Additionally, we also design the iterative-version of PhaseAT. Due to the space limit, we include the details in Supplementary material.}

\begin{algorithm}[t] \caption{Phase-shifted Adversarial Training} \label{algo}
\begin{algorithmic}[1]
\Require Training epochs $T$, dataset $\mathcal{D}$, dataset size $N$, perturbation size $\epsilon$, perturbation step $\alpha$, networks $\mathcal{T}$
\For {$t= 1$ $...$ $T$} 
    % \For {$j= 1$ $...$ $N/$}
     \ForEach{ Batch $(X, Y) \sim \mathcal{D}$  }
        \State $\Delta = $ Uniform$(-\epsilon, \epsilon)$
        \State \textcolor{gray}{// Alternate training across mini-batches}
        \If {$j$ \% 2 == 0} 
        \State $\Delta = \Delta + \alpha \cdot$ sign($\nabla_{\Delta}\ell_{\text{ce}}(\mathcal{T}(X + \Delta),Y)$)
        \Else
        \State $\Delta = \Delta + \alpha \cdot$ sign($\nabla_{\Delta}\ell_{\text{ce}}(\mathcal{T}_{0}(X + \Delta),Y)$)
        \EndIf
        \State $\Delta = $ max(min($\Delta, \epsilon$), $-\epsilon$)
        
        \State \textcolor{gray}{// Sampling frequencies for $\mathcal{T}$ (Eq.~\eqref{eq:freq_select1}, \eqref{eq:freq_select2}})
        \State ${d} = \left\vert \mathcal{F}(X) - \mathcal{F}(X+\Delta)\right\vert$
        \State $\{\omega_m\}_{m=1}^{M} \sim \text{Multinomial}(d)$
        \State \textcolor{gray}{// Optimize networks based on loss (Eq.~\eqref{eq:objective}}
        \State $\theta = \theta - \nabla_{\theta}[\ell_{\text{adv}}(X, Y)$)]
        % \EndFor
    \EndFor
\EndFor
% Fourier updates 
% Frequency selection 
\end{algorithmic}
\label{alg:main}
\end{algorithm}

% \begin{algorithm}[t] \caption{Phase-shifted Adversarial Training} \label{algo}
% \begin{algorithmic}[1]
% \Require Training epochs $T$, Dataset size $N$, Perturbation size $\epsilon$, Perturbation step $\alpha$, Trainable networks $\mathcal{T}$, Cosine similarity function $CS(\cdot, \cdot)$
% \For {$t= 1$ $...$ $T$} 
%     \For {$j= 1$ $...$ $N$}
%         \State $\delta = $ Uniform$(-\epsilon, \epsilon)$
%         \If {$j$ \% 2 == 0} \Comment{Alternate training}
%         \State $\delta = \delta + \alpha \cdot$ sign($\nabla_{\delta}\ell(\mathcal{T}(x_j + \delta),y_j)$)
%         \Else
%         \State $\delta = \delta + \alpha \cdot$ sign($\nabla_{\delta}\ell(\mathcal{T}_{0}(x_j + \delta),y_j)$)
%         \EndIf
%         \State $\delta = $ max(min($\delta, \epsilon$), $-\epsilon$) 
%         \State $\theta = \theta - \nabla_{\theta}[\ell(\mathcal{T}(x_j + \delta), y_j)$ + $CS(\mathcal{T}(x_j + \delta), \mathcal{T}_{0}(x_j + \delta)$)]
%     \EndFor
% \EndFor
% % Fourier updates 
% % Frequency selection 
% \end{algorithmic}
% \label{alg:main}
% \end{algorithm}

The FGSM attack is formulated by initializing the perturbation on uniform distribution ranging ($-\epsilon$, $\epsilon$). The perturbation is then updated based on the sign of the gradients of the cross-entropy with the ground-truth labels. Since PhaseDNN selects the frequencies in a stochastic manner, the generated attack can encourage PhaseDNN to better fit the diverse components having high-frequencies (detailed in Section \ref{sect:white}). Moreover, we perform the phase-shift in alternate mini-batches which further diversifies the attacks used for adversarial training similar to the previous studies \citep{sriramanan2020guided,sriramanan2021towards}.

 \paragraph{Adversarial training.} We improve the model's robustness against generated attacks by replacing the clean inputs with perturbed ones during training. Additional improvements can be achieved by regularization to minimize the effects of potential white-box attacks. One possible white-box attack would set all shift frequencies to zero, resulting in gradients similar to normal AT rather than PhaseAT. Our regularization term encourages the model to behave differently than the standard AT. We implement this by minimizing the prediction similarity between the phase-shifted model and the model that does not have the phase-shift. In summary, the total objective function is the sum of the cross-entropy loss and the regularization.
\begin{equation}\label{eq:objective}
\begin{aligned}
    \ell_{adv}(x, y) &=  \ell_{ce}(\mathcal{T}(x+\delta), y) \text{ }+ \\& \left\vert\frac{\sigma(\mathcal{T}(x+\delta)) \cdot \sigma(\mathcal{T}_{0}(x+\delta))}{\Vert \sigma(\mathcal{T}(x+\delta)) \Vert \Vert \sigma(\mathcal{T}_{0}(x+\delta)) \Vert } \right\vert
\end{aligned}    
\end{equation}
where $\ell_{ce}$ is the cross entropy function, $\sigma$ indicates softmax function, $\mathcal{T}_{0}$ indicates the model without the phase-shift term (i.e., $\omega_m = 0,$ $ \forall$ $m \in [1,M]$). By encouraging the model to have different predictions than normal AT, the model achieves robustness against the normal AT attack as well \footnote{We provide the ablation study about the regularization term in Supplementary material.} The proposed defense is detailed in Algorithm-\ref{alg:main}.

\section{Evaluations}\label{sect:exp}
% In this section, we evaluate the proposed method on benchmark settings of the adversarial training.
\subsection{Experimental Setup}
\paragraph{Baselines.}
% Compared method 
We compare our method with both non-iterative and iterative methods. Non-iterative methods updates the perturbation once to generate the attacks, whereas iterative methods generate the perturbations through multiple optimization steps. We choose three recent non-iterative methods for comparison, namely FBF \citep{wong2019fast}, GAT\citep{sriramanan2020guided}, and NuAT \citep{sriramanan2021towards}. For iterative methods, we select four representative baselines, i.e., PGD \citep{madry2017towards}, TRADES \citep{zhang2019theoretically}, GAIRAT \citep{zhang2020geometry}, and AWP \citep{wu2020adversarial}. We carefully set the hyper-parameters for the aforementioned methods. We use ResNet-18 \citep{he2016deep} and WideResNet-34-10 \citep{zagoruyko2016wide} architectures \citep{he2016deep} for the evaluations and, the detailed settings are listed in Supplementary material.

\paragraph{Datasets and Evaluations.}
Following recent works, we perform experiments on two datasets: CIFAR-10 \citep{krizhevsky2009learning} and ImageNet \citep{deng2009imagenet}. To compare with the recent works, we use ImageNet-100 \citep{sriramanan2020guided,sriramanan2021towards} which is the small version of ImageNet including 100 classes. To evaluate adversarial robustness, we consider both white-box and black-box attacks. For white-box attacks, PGD and AutoAttack (AA) are mainly used, each of which contains $L_{\infty}$ adversarial constraint of $\epsilon = 8/255$. Specifically, the AA attacks involve APGD$_{ce}$, APGD$_{dlr}$, APGD$_{t}$ \citep{croce2020reliable} and FAB$_{t}$ \citep{croce2020minimally}\footnote{We include the details of each attack in Supplementary material.}. 
% \textcolor{red}
% {The black-box attacks include transfer-based and query-based attacks. For the transfer-based attack, we attack each baseline by generating PGD-7 perturbations of other networks (e.g., VGG-11 and ResNet-18 that is differently initialized). For query-based attacks, we use square attacks \citep{andriushchenko2020square} with 5,000 query budgets. The results against black-box attacks are included in Supplementary material.}
The black-box attacks include transfer-based and query-based attacks. For the transfer-based attack, we generate PGD-7 perturbations using other networks (e.g., VGG-11 and ResNet-18 that is differently initialized) to attack each baseline. For query-based attacks, we use square attacks~\citep{andriushchenko2020square} with 5,000 query budgets. The results of black-box attacks are included in Supplementary material.

% \textcolor{red}{\textbf{Adaptive Attack.}
% For reliable estimation, we carefully design adaptive attacks for the proposed defenses to adversarial examples. We discuss the adaptive attacks in Section \ref{sup:adaptive} in Supplementary material and apply adaptive attacks to all experiments. For example, due to the process of multinomial sampling for the frequencies, we use the expectation-over-transformation (EOT) \citep{athalye2018synthesizing} for the white-box attacks (i.e., PGD, AA).}

% \noindent\textbf{Adaptive Attack.}
% For reliable evaluation of the proposed adversarial training, we carefully design adaptive attacks for the proposed defense to adversarial examples, which are used in all experiments. Specifically, we use the expectation-over-transformation (EOT)~\citep{athalye2018synthesizing} as a white-box attack because the proposed method samples frequencies from a multinomial distribution. The details of the adaptive attacks are detailed in the supplementary material.

\begin{table}[h]
\centering
\caption{Performance evaluation of the different designs of adaptive attacks on CIFAR-10 dataset.}\label{wrap-tab:1}
\begin{adjustbox}{max width=\textwidth}
\begin{tabular}{@{}lcc@{}}\\\toprule  
Attack  & Accuracy \\\midrule 
PGD$_{50}$                      &       61.5  \\  \midrule
PGD$_{50}$ + EOT                &       60.9 \\  
PGD$_{50}$ + Frequency          &       59.7  \\  
PGD$_{50}$ + EOT + Frequency    &       59.5  \\  \bottomrule
\end{tabular}
\end{adjustbox}
\label{tab:adaptive_attack}
% \end{wraptable} 
\end{table}

\begin{table*}[t]
    \caption{Performance evaluation on CIFAR-10 dataset against white-box attacks on two different architectures. Best and second best results are highlighted in boldface and underline, respectively.}
    \begin{subtable}{.5\linewidth}
      \centering
    \begin{tabular}{@{}ccccc@{}}
    \toprule
     Method & Clean accuracy  & PGD$_{50}$\tiny{+EOT} & AA\tiny{+EOT} \\ \midrule
     Normal & 93.1   & 0.0    & 0.0  \\ \midrule
     FBF  & \underline{84.0}  & 43.8     & 41.0  \\
     GAT   & 80.5    & 53.2    & 47.4   \\
     NuAT    & 81.6    & 52.0     & 48.3   \\
     PhaseAT (ours)  & \textbf{86.2}   & \textbf{59.5}    & \textbf{52.1}  \\ \midrule \midrule
                                    
     PGD-AT   & 81.3  & 50.8     & 47.3   \\
     TRADES   & 79.5   & 52.2   & 48.5 \\
     GAIRAT   & 82.7   & \underline{56.3}   & {34.5} \\
     AWP   & 81.8    & {54.8}    & \underline{51.1}  \\ \bottomrule
    \end{tabular}
    \caption{ResNet-18}
    \end{subtable}
    \begin{subtable}{.5\linewidth}
      \centering
        \begin{tabular}{@{}ccccc@{}}
        \toprule
         Method & Clean accuracy  & PGD$_{50}$\tiny{+EOT} & AA\tiny{+EOT} \\ \midrule
         Normal & 94.5   & 0.0     & 0.0   \\ \midrule
         FBF    & 82.1   & 54.4     & 51.3   \\
         GAT    & 84.7    & 56.1     & 52.1   \\
         NuAT    & 85.1    & 54.6    & 53.1   \\
         PhaseAT (ours)  & \textbf{88.8}   & \textbf{62.3}   & \textbf{59.2}  \\ \midrule \midrule
                                        
         PGD-AT   & 86.7    & 56.8     & 54.7   \\
         TRADES   & \underline{87.6}    & 55.9     & 53.9   \\
         GAIRAT   & 85.3    &  57.6    & 42.6   \\
         AWP      & 86.9    & \underline{60.4}     & \underline{56.5}   \\ \bottomrule
        \end{tabular}
        \caption{WideResNet-34-10}
    \end{subtable} 
    \label{exp:cifar10}
% \vspace{-1cm}
\end{table*}

\begin{figure*}[t]
\centering
\subfloat[Low-frequency errors]{%
  \includegraphics[width=0.46\linewidth]{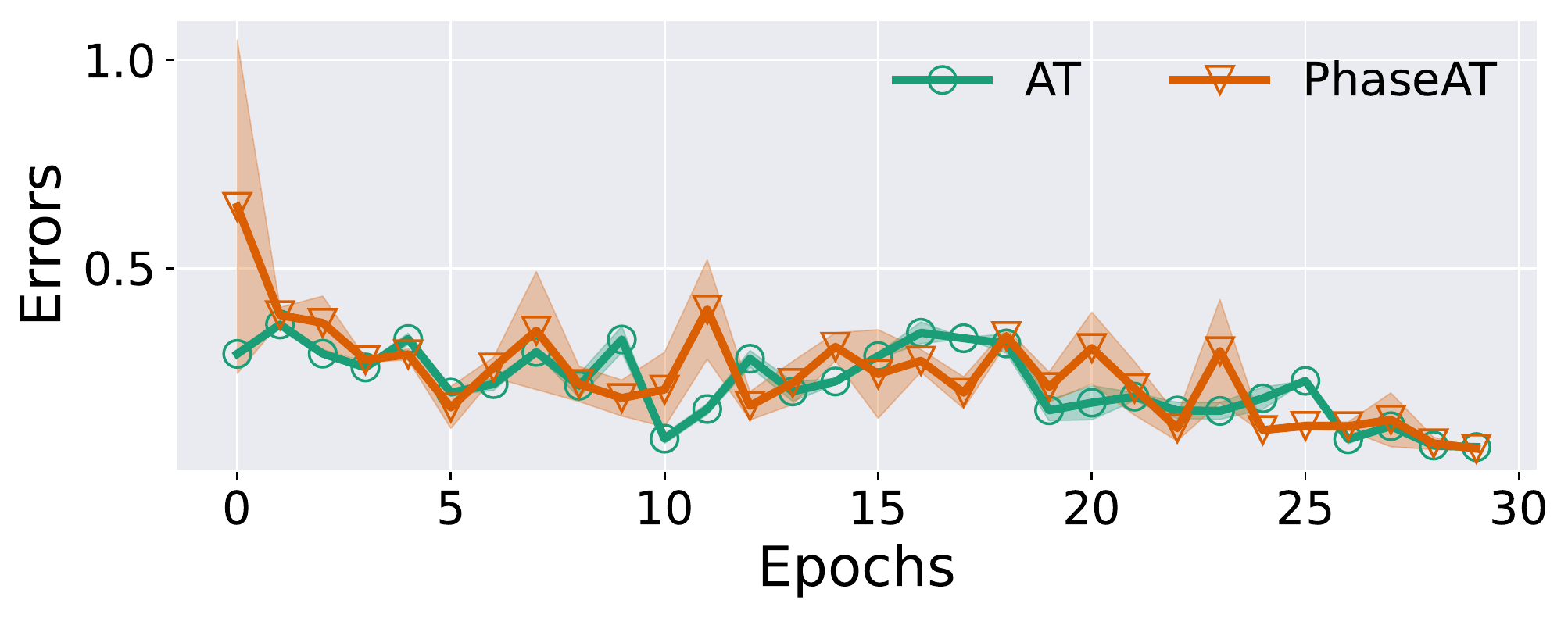}%
}
\subfloat[High-frequency errors]{%
  \includegraphics[width=0.46\linewidth]{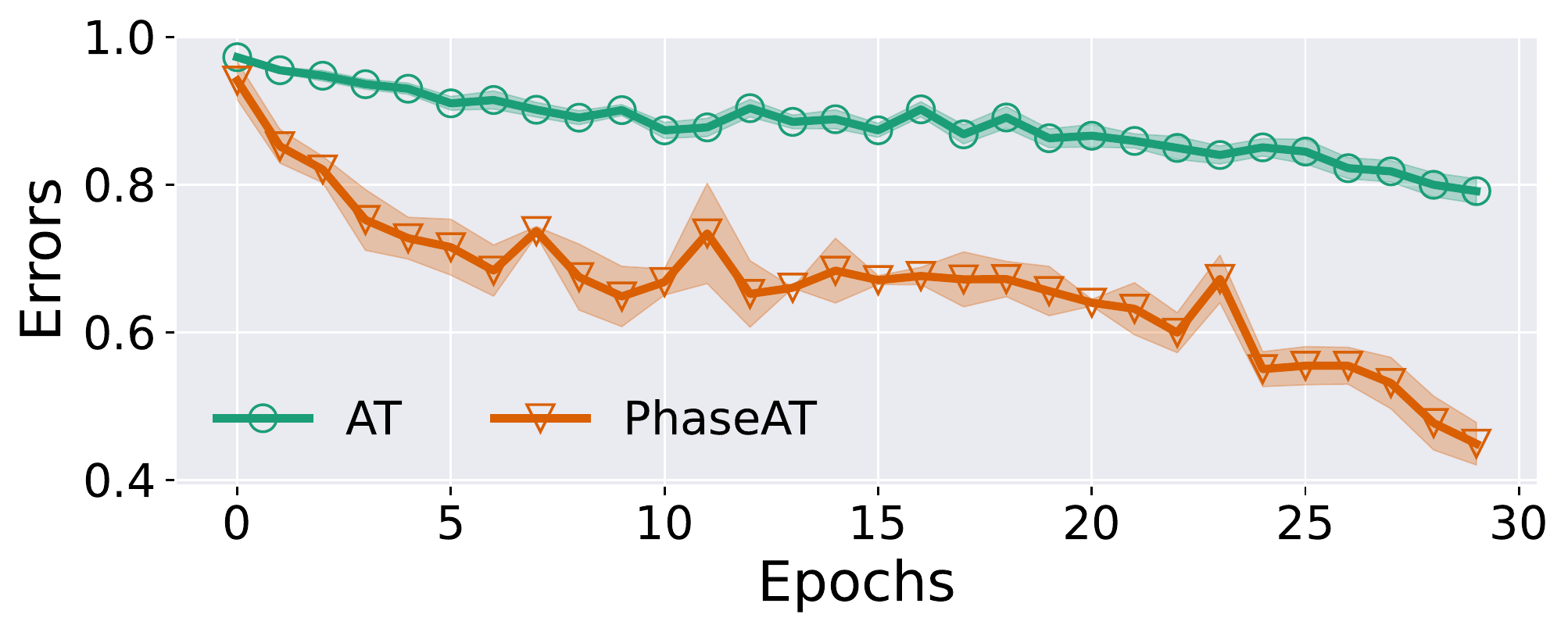}%
}
\caption{Errors of frequency components (high, low) between the training dataset and neural networks for normal adversarial training method and PhaseAT.}
\label{fig:freq_cifar10}
\end{figure*}

\paragraph{Consideration of Adaptive Attack.}

% \begin{wraptable}{r}{6cm}

Compared to previous methods, PhaseAT includes stochastic process to learn diverse frequencies of the datasets, which prevents reliable evaluation of the adversarial robustness. Following the guideline \citep{tramer2018ensemble}, we carefully design the adaptive attack for our method. First, to circumvent the stochastic process, the adaptive attack involves the expectation over transformation (EOT) \citep{athalye2018synthesizing} to approximate the true gradient of the inference model. Second, we assume that the adversary has full knowledge of the learning algorithm. Hence the adversary could mimic the strategy of frequency selection used in the training/inference phase, i.e., multinomial distribution of the discrepancy. We tabulate how these adaptive attacks affect our method in Table \ref{tab:adaptive_attack}. 

This shows that reducing the effect of the stochastic properties by EOT decreases the performance of our method. Particularly, mimicking the frequency selection strategy further reduces the performance. However, note that PhaseAT still reveals strong adversarial robustness even though the adaptive attacks are applied (refer to main paper), which requires ten times more costs to generate the attacks compared to other baselines. Such adaptive attacks are applied to our method for a fair comparison.

\subsection{Main Results against White-box Attacks}\label{sect:white}

Before comparing the performance of all methods, we first verify whether our method learns high-frequency components faster than the model without the phase-shift. To that end, we use the filtering method \citep{xu2019frequency} used in previous analysis (Figure \ref{fig:motivation}). Figure \ref{fig:freq_cifar10} shows the errors for each frequency. Here, we use the settings of CIFAR-10. For the low-frequency part, the errors between PhaseAT and AT are negligible, whereas the errors of the high-frequency between two methods is noticeable, demonstrating our hypothesis that PhaseAT can efficiently learn high-frequency components.  

We then confirm that fast learning of high-frequency components leads to better robustness. Table \ref{exp:cifar10} shows the comparison results. Here, we use the two different networks, ResNet-18 and WideResNet-34-10, to verify the generality of PhaseAT. Non-iterative methods tend to show lower accuracy than iterative methods. However, the non-iterative version of PhaseAT outperforms iterative baselines in terms of both standard and robust accuracy. For example, PhaseAT shows 5.3\% and 8.5\% performance improvement over AWP in terms of standard and PGD accuracy, respectively, and comparable performance for AA.  In the experimental results, PhaseAT outperforms FBF, the major distinction of which is the phase-shift functionality. This suggests that learning high-frequency information properly is particularly beneficial for learning adversarial data. 

We turn our focus to a larger dataset, i.e., ImageNet. We compare our method with all non-iterative methods and tabulate the results in Table \ref{exp:imagenet}. We again find a similar trend in performance to that of CIFAR-10 except for the strongest baseline is the non-iterative method (i.e., NuAT). While the proposed method shows the comparable performance for clean accuracy, it outperforms others by a large margin in terms of the robust accuracy against the AA attack. This shows that PhaseAT can be scalable to larger datasets. 

\begin{table}[t]
\centering
\caption{Performance evaluation on ImageNet-100 dataset against white-box attacks. Best and second best results are highlighted in boldface and underline, respectively. The results for each baseline come from the previous works \citep{sriramanan2020guided,sriramanan2021towards}.}
\begin{adjustbox}{max width=\textwidth}
\begin{tabular}{@{}cccc@{}}
\toprule
 Method & Clean accuracy  & AA\tiny{+EOT} \\ \midrule
 Normal & 77.1    & 0.0  \\ \midrule
 FBF   & \textbf{70.5}   & 20.4  \\
 GAT  & 68.0    & 28.9  \\
 NuAT  & 69.0   & 32.0  \\
 PhaseAT (ours)  & \underline{{69.2}}    & \textbf{35.6}  \\ \midrule \midrule
 PGD-AT   & 68.6    & \underline{33.0}  \\
 TRADES  & 62.9  & 31.7  \\
 AWP   & 64.8    & 29.2  \\ \bottomrule
\end{tabular}
\end{adjustbox}
\label{exp:imagenet}
% \vspace{-0.5cm}
\end{table}

% \begin{table}[t]
% \centering
% \caption{Performance evaluation on CIFAR-10 dataset against two different black-box attacks.}
% \begin{tabular}{@{}lcccc@{}}
% \toprule
% \multirow{2}{*}{Method} & \multirow{2}{*}{clean accuracy} & \multicolumn{2}{c}{Transfer-based attack} & \multirow{2}{*}{Score-based attack} \\ \cmidrule(lr){3-4}
%                   &                                    & VGG-11                 & ResNet-18            &                                     \\ \midrule
% FBF \citep{wong2019fast} & 84.0       & 80.5   & 80.6           & 53.5      \\
% GAT \citep{sriramanan2020guided}   & 80.5             & 79.8   & 80.3            & 54.1      \\
% NuAT \citep{sriramanan2021towards}   & 81.6             & 79.5   & 80.5            & 56.7      \\
% PhaseAT (Ours.)             & 86.2             & 83.8   & 85.0           & 76.5      \\  \midrule \midrule
% RND \citep{qin2021random}             & 93.0             & -  & -          & 82.9      \\  \bottomrule
% \end{tabular}
% \label{exp:black_cifar}
% \end{table}

\begin{figure}[t]
\centering
\subfloat[Clean accuracy]{%
  \includegraphics[width=0.95\linewidth]{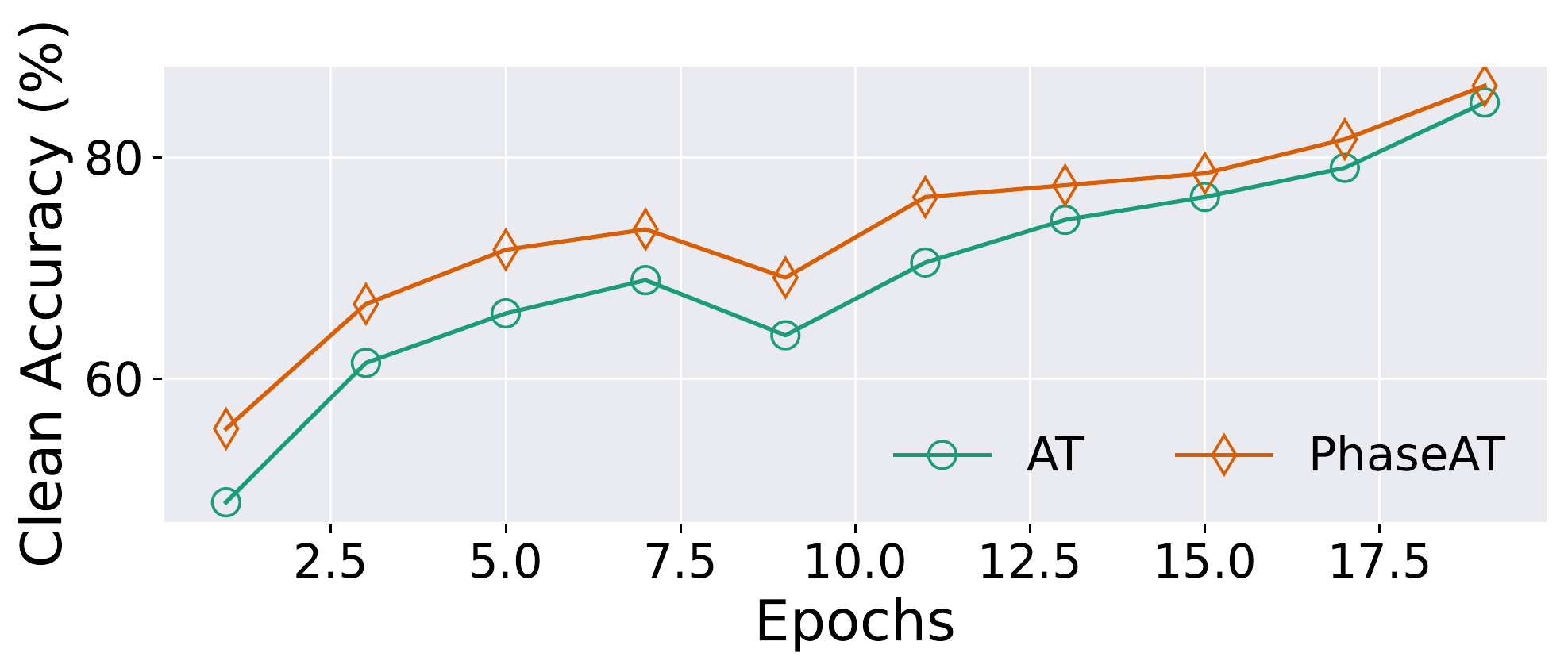}%
}
% \vspace{0.3cm}

\subfloat[Robustness Accuracy]{%
  \includegraphics[width=0.95\linewidth]{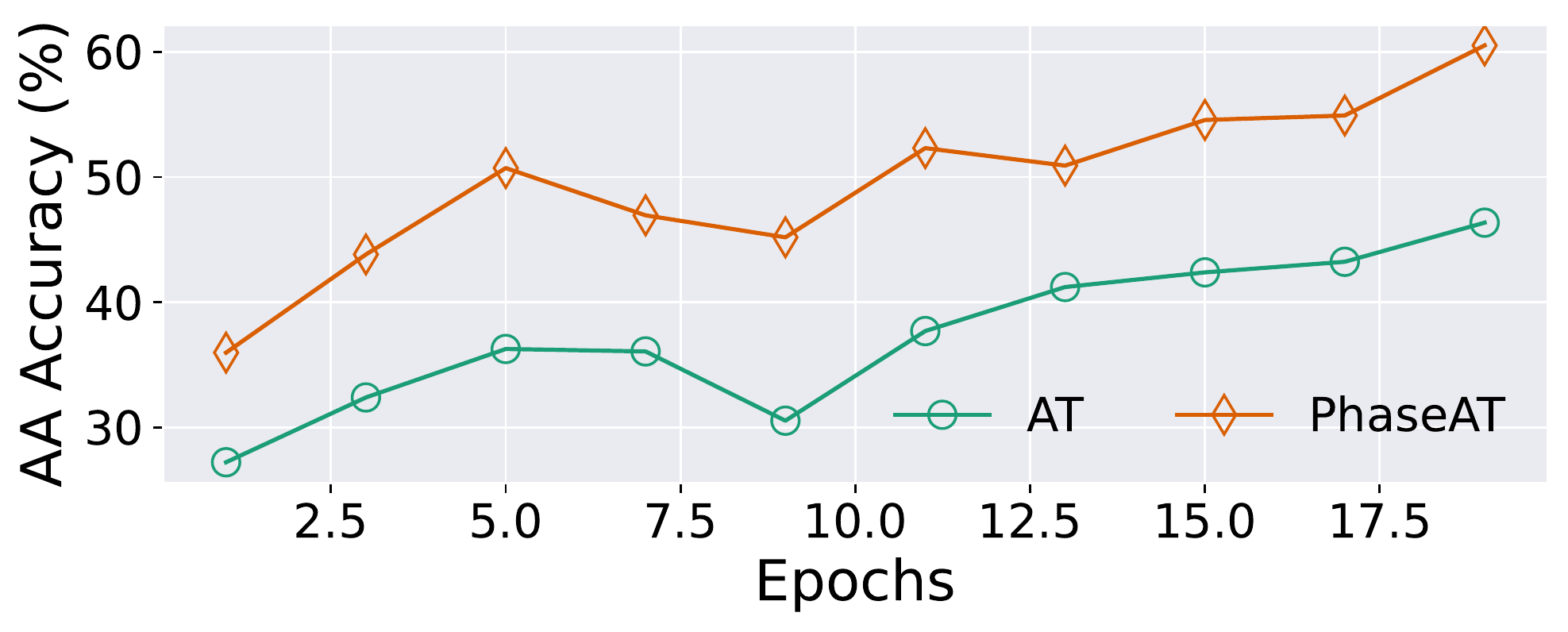}%
}
\caption{Clean and robust accuracy during the training.}
% This shows that PhaseAT has better and faster convergence to both standard and robust accuracy.
\label{fig:curv}
\end{figure}

\begin{figure}[t]
\centering
\subfloat[Sensitivity to head]{%
  \includegraphics[width=0.95\linewidth]{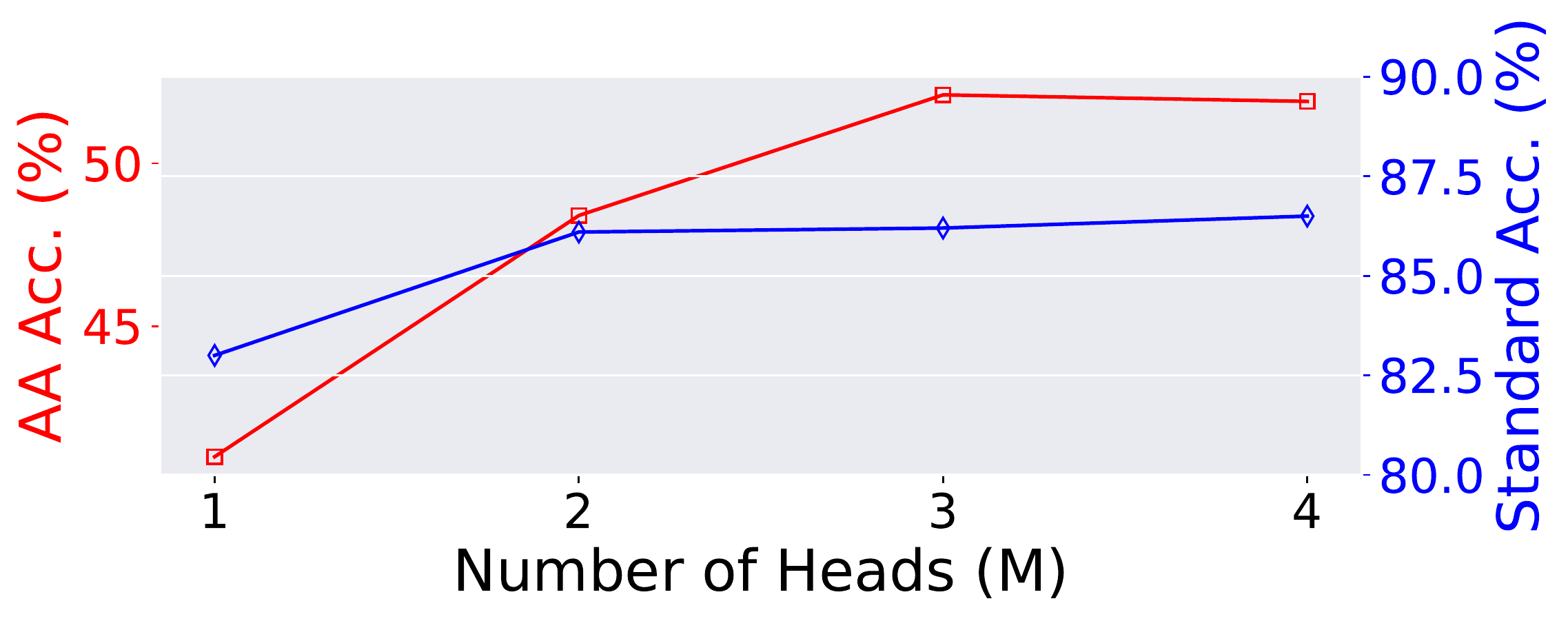}%
}

\subfloat[Sensitivity to frequency]{%
  \includegraphics[width=0.95\linewidth]{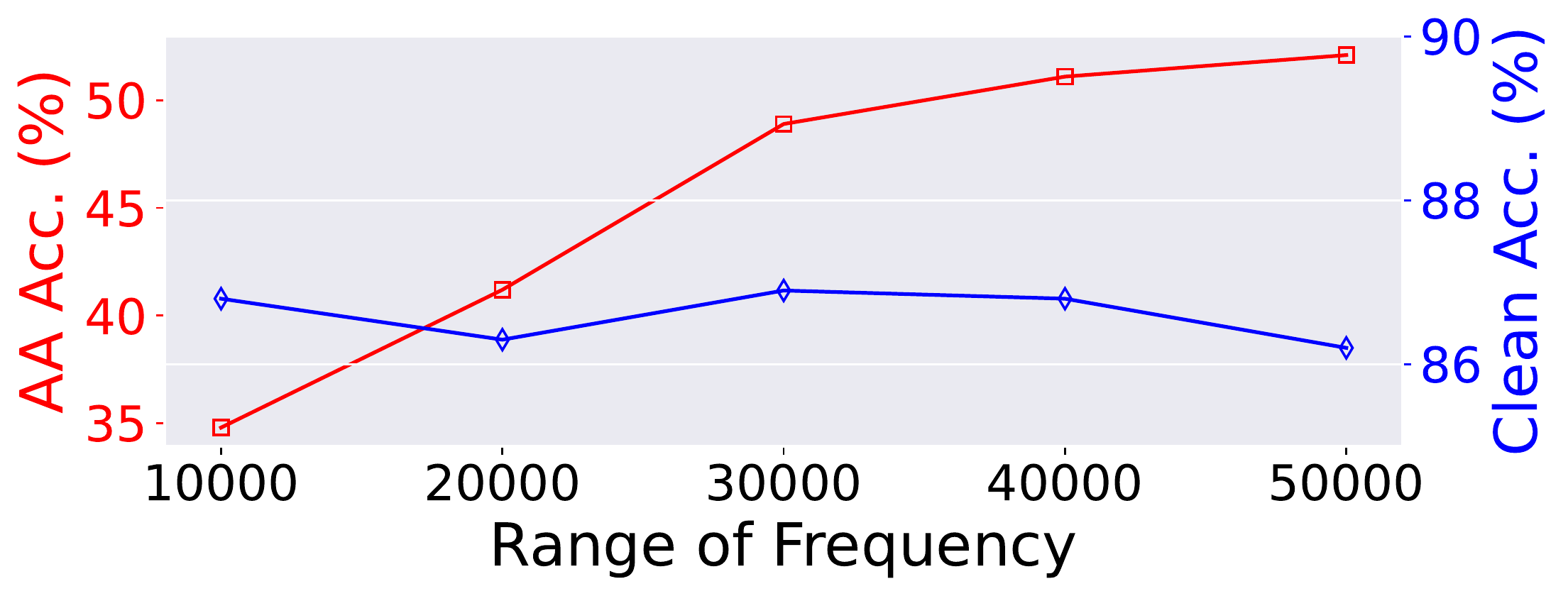}%
}
\caption{Sensitivity to the different number of heads and the range of frequency for PhaseAT.}
\label{fig:ablation}
\end{figure}

\subsection{Convergence of Clean and Robust Accuracy}
We verify that faster learning on high-frequency information leads to better convergence for robust accuracy. To that end, we report the standard and AA accuracy during each training epoch. We use the same settings of CIFAR-10 and tabulate the results on Figure \ref{fig:curv}. Compared to the normal adversarial training (denoted as AT), PhaseAT shows faster and better convergence both on standard and robust accuracy. This indicates that the training model has smoothed predictions near the data, effectively reducing the oscillating property of predictions when adding adversarial perturbations.
 
\subsection{Sensitivity to the Different Settings of PhaseAT}
We analyze the sensitivity of PhaseAT on different settings. Since the most significant components of PhaseAT are the head and frequency, we mainly control the number of heads (i.e., $M$) and the range of frequency. Figure \ref{fig:ablation} shows how the change of each parameter affects the standard and robust accuracy on CIFAR-10. We first see that utilizing more heads leads to the improved accuracy while incurring additional computation costs. In addition, the robust accuracy is more sensitive to the different number of heads, whereas the clean accuracy does not differ significantly when more than two heads are used. When it comes to the frequency range, we observe that utilizing the wider range of the frequency leads to the improved robust accuracy. For the clean accuracy case, on the other hand, the wider range has no significant impact, which is consistent with the previous finding that low-frequency typically dominates in the real dataset \citep{xu2019frequency}. In the experiments, we use the three heads and 50k of the frequency range based on the above validation results. For the frequency range case, no further improvement was observed for more than the 50k range.

\subsection{Computational Complexity}

\begin{figure}[t]
\centering
\includegraphics[width=0.95\linewidth]{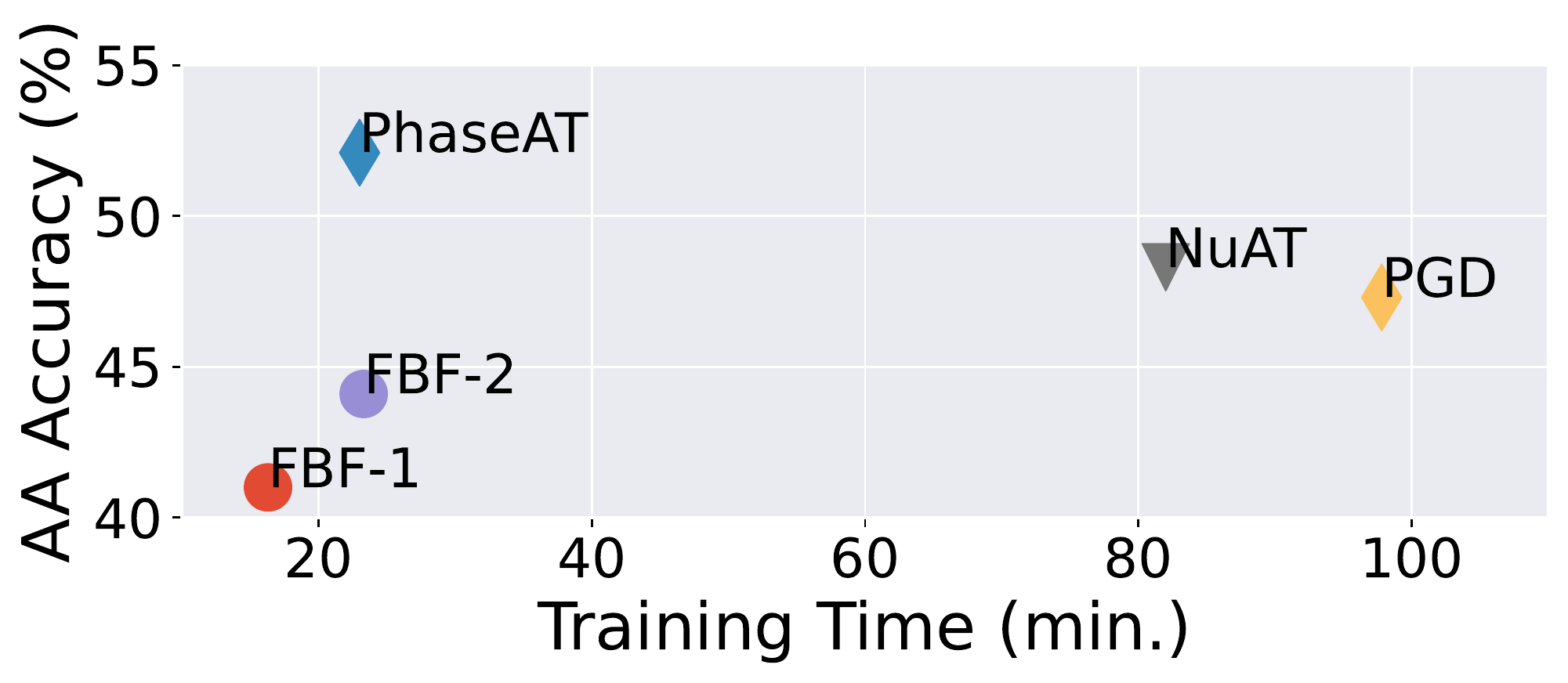}
\caption{Comparison of the training time of each method to achieve the best robust accuracy.}
\label{fig:train_time}
\end{figure}

We compare the training time required to achieve the best robust accuracy for each method. To fairly measure it, we use the optimized schedule in the implementations (NuAT\footnote{\url{https://github.com/locuslab/fast_adversarial}. }, FBF and PGD\footnote{\url{https://github.com/val-iisc/NuAT}}). Figure \ref{fig:train_time} plots the comparison on CIFAR-10. The fastest defense amongst all is FBF-1 and PhaseAT shows the comparable training time. This indicates the additional time required by Fourier transform and forward-propagation to multiple heads is negligibly small. We also compare the different versions of FBF, namely FBF-2, which uses more training epochs than FBF-1 to match the training time with PhaseAT. Despite FBF's increased accuracy (41.0 to 44.1\%), PhaseAT outperforms it by a significant margin, demonstrating the efficacy of the proposed approach.

\section{Related Work}

\subsection{Adversarial Training}

One of the most effective defenses against adversarial attacks is adversarial training. Specifically, iteratively updating the attacks during training tends to show better robustness as the adversary typically performs multiple updates to generate stronger attacks \citep{madry2017towards,zhang2019theoretically,wu2020adversarial}. However, the adversarial robustness comes at a large computational cost and highly increased training time due to the multiple optimizations. Hence the research towards non-iterative methods has been getting attention \citep{wong2019fast,sriramanan2020guided,sriramanan2021towards}. The common strategy is to make the FGSM-based training robust against iterative attacks. For example, \citep{wong2019fast} discover that training with FGSM-based training can be a stronger defense against iterative attacks. This FGSM can be achieved by initializing the perturbation as uniform distribution with properly adjusted perturbation and step sizes. Instead of the initialization, \citep{sriramanan2020guided} propose a non-iterative adversarial training method by introducing $L_{2}$ regularization for smoothing the loss surface initially and further reducing the weight of this term gradually over iterations for better optimization. Similarly, \citep{sriramanan2021towards} enhance the above relaxation term by involving the use of nuclear norm to generate a more reliable gradient direction by smoothing the loss surface.

\subsection{F-Principle in Standard Training}
The F-principle was first discovered empirically in \citep{rahaman2019spectral} and \citep{xu2019training} independently.
A theoretical proof was given by \citep{xu2019frequency} only for a one-hidden layer $\tanh$ network with a mean-squared error loss function. It was later extended in \citep{luo2019theory} to multi-layer networks with general activation functions and general loss functions excluding cross-entropy loss function.

\section{Conclusion}\label{sect:conclusion}
% Response Frequency
% Observation
% Redesign PhaseDNN
% Theorem
% Performance improvement
In this work, we have analyzed the behavior of adversarial training through the lens of the response frequency and have empirically observed that adversarial training causes neural networks to have slow convergence on high-frequency components. To recover such slow convergence, we first prove that F-principle still holds in adversarial training involving more generalized settings. On this basis, we propose phase-shifted adversarial training, denoted as PhaseAT, in which the model learns high-frequency contents by shifting them to low-frequency ranges. In our method, we redesign the PhaseDNN to adopt high-dimensional inputs and effectively learn adversarial data. For reliable evaluation, we carefully design adaptive attacks and evaluate the proposed method on two popular datasets, namely CIFAR-10 and ImageNet. The results clearly show that PhaseAT shows strong performance compared to other baselines, demonstrating the efficacy of faster high-frequency learning in adversarial training. While PhaseAT enables faster training on adversarial examples, the pre-processing phase to derive the first principal component can be overhead as the number of datasets increases. In future work, we will explore the way to derive the first principle component efficiently.

\section*{Acknowledgment} 
This work was supported by the Center for Advanced Computation and a KIAS Individual Grant (MG082901) at Korea Institute for Advanced Study and the POSCO Science Fellowship of POSCO TJ Park Foundation (S. Kim), and by NRF-2022R1A2C1011312 (I. Seo).

% References
\bibliography{kim_707}

\onecolumn

\section{Supplementary Materials}

\subsection{Filtering Method for Frequency Analysis}

Motivated by the examination of F-principle \citep{xu2020frequency}, we use the filtering method to analyze the behavior of the neural networks in adversarial training. The idea is to split the frequency domain into two parts, i.e., low-frequency and high-frequency parts. However, the Fourier transform for high-dimensional data requires high computational costs and large memory footprints. As an alternative, we use the Fourier transform of a Gaussian function $\hat{G}$.

Let the original dataset be $\{x_j, y_j\}_{j=0}^{N-1}$, and the network output for $x_j$ be $\mathcal{T}_j$. The low frequency part of the training dataset can be derived by
\begin{equation}
    y_{j}^{low, \delta} = \frac{1}{C_j} \sum_{m=0}^{N-1}y_{m} G^{\delta}(x_j - x_m)
\end{equation}
where $C_j = \sum_{m=0}^{N-1}G^{\delta}(x_j - x_m)$ is a normalization factor, and $\delta$ is the variance of the Gaussian function (we fix $\delta$ to 3). The Gaussian function can be represented as 
\begin{equation}
    G^{\delta}(x_j - x_m) = \exp(-\vert x_j - x_m\vert^2/(2\delta)).
\end{equation}
Then, the high-frequency part can be derived by $y_{j}^{high, \delta} \delequal y_j - y_{j}^{low, \delta}$. We also compute the frequency components for the networks, i.e, $\mathcal{T}_{j}^{low, \delta}, \mathcal{T}_{j}^{high, \delta}$ by replacing $y_j$ with the outputs of networks, i.e., $\mathcal{T}_j$. Lastly, we calculate the errors to quantify the convergence in terms of low- and high-frequency.
\begin{equation}
    e_{low} = \left(\frac{\sum_{j}\left\vert y_{j}^{low,\delta} - \mathcal{T}_j^{low,\delta} \right\vert^2}{\sum_{j} \left\vert y_{j}^{low,\delta}\right\vert^2} \right)^{\frac{1}{2}}
\end{equation}

\begin{equation}
    e_{high} = \left(\frac{\sum_{j}\left\vert y_{j}^{high,\delta} - \mathcal{T}_j^{high,\delta} \right\vert^2}{\sum_{j} \left\vert y_{j}^{high,\delta}\right\vert^2} \right)^{\frac{1}{2}}
\end{equation}

\subsection{Iterative-version of PhaseAT}

\begin{algorithm}[h] \caption{Phase-shifted Adversarial Training (Iterative version)} \label{algo}
\begin{algorithmic}[1]
\Require Training epochs $T$, Dataset size $N$, PGD steps $P$, Perturbation size $\epsilon$, Perturbation step $\alpha$, Trainable networks $\mathcal{T}$, Cosine similarity function $CS(\cdot, \cdot)$
\For {$t= 1$ $...$ $T$} 
    \For {$j= 1$ $...$ $N$}
        \State $\delta = $ Uniform$(-\epsilon, \epsilon)$
        \For {$k= 1$ $...$ $P$} \Comment{Multiple updates of perturbations}
            \If {$j$ \% 2 == 0} \Comment{Alternate training on mini-batches }
            \State $\delta = \delta + \alpha \cdot$ sign($\nabla_{\delta}\ell(\mathcal{T}(x_j + \delta),y_j)$)
            \Else
            \State $\delta = \delta + \alpha \cdot$ sign($\nabla_{\delta}\ell(\mathcal{T}_{0}(x_j + \delta),y_j)$)
            \EndIf
            \State $\delta = $ max(min($\delta, \epsilon$), $-\epsilon$) 
        \EndFor
        \State $\theta = \theta - \nabla_{\theta}[\ell(\mathcal{T}(x_j + \delta), y_j)$ + $CS(\mathcal{T}(x_j + \delta), \mathcal{T}_{0}(x_j + \delta)$)]
    \EndFor
\EndFor
% Fourier updates 
% Frequency selection 
\end{algorithmic}
\label{alg:main2}
\end{algorithm}

For training efficiency, we design PhaseAT as a non-iterative method based on the FGSM perturbation \citep{wong2019fast}. To confirm the effect of stronger attacks in the training process of PhaseAT, we additionally introduce an iterative version of PhaseAT. Since PhaseAT is not closely related to the perturbation generation, we replace the FGSM perturbation with the perturbation generated from PGD \citep{madry2017towards}. The overall algorithm is shown in Algorithm-\ref{alg:main2}.

\subsection{Details about Evaluation}
\subsubsection{Attack Configuration}
In our work, we mainly adopt the projected gradient descent (PGD) \citep{madry2017towards} and auto-attack (AA) \citep{croce2020reliable} to evaluate baselines. PGD is constructed by multiple updates of adversarial perturbations, and AA is the ensemble of strong attacks including the variants of PGD. Typically, AA is considered one of the strongest attacks. The details about each attack of AA are as follows:
\begin{itemize}

    \item Auto-PGD (APGD) \citep{croce2020reliable}: This is parameter-free adversarial attack that adaptively changes the step size by considering the optimization of the perturbations. APGD has three variations depending on loss functions: APGD$_{ce}$, APGD$_{dlr}$, and APGD$_t$\footnote{Subscript $ce$ and $dlr$ on APGD indicates the \textit{cross-entropy loss} and \textit{difference of logits ratio}, respectively, and $t$ stands for targeted attacks. The attacks without $t$ subscripts are non-targeted attacks.}.
    
    \item FAB \citep{croce2020minimally}: This attack minimizes the norm of the perturbation necessary to achieve a misclassification. FAB has two variants, FAB and FAB$_{t}$.
    
    \item Square \citep{andriushchenko2020square}: Compared to others, this attack belongs to the black-box attacks and is also known as score-based attack. This attack iteratively inserts an artificial square to the inputs to search optimal perturbations causing huge changes on predictions.
    
\end{itemize}
We set the hyper-parameter settings for each attack based on \textit{standard} version of AA in \textit{robust-bench} framework \citep{croce2021robustbench}. Note that we exclude Square attack from the AA because the stochastic process in PhaseAT can be robust against Square \citep{qin2021random}, which prevents the fair comparison with other baselines which do not include stochastic process. We thus move the results of Square attack to the Supplementary Section \ref{sect:black}.

\subsubsection{Dataset Information}
We evaluate each baseline on two benchmark datasets, CIFAR-10 and ImageNet. CIFAR-10 \citep{krizhevsky2009learning} consist of 60,000 images of 32×32×3 size for 10 classes, and ImageNet contains 1.2M images of 224×224×3 size for 1,000 classes. Instead of existing ImageNet, we use the smaller version of ImageNet which used in recent baselines \citep{sriramanan2020guided,sriramanan2021towards}, which contains 120K images of 224×224×3 size for 100 classes \footnote{Selected classes are listed in \url{https://github.com/val-iisc/GAMA-GAT}}.

\subsubsection{Baseline Setting}
PhaseAT is compared to both non-iterative (FBF, GAT, and NuAT) and iterative (FBF, GAT, and NuAT) methods (PGD, TRADES, and AWP). The hyper-parameter settings of each baseline are listed in Table \ref{tab:param}. Since the evaluation results on ImageNet come from previous works \citep{sriramanan2020guided,sriramanan2021towards}, the table only includes the parameters reported in these works (unknown parameters are denoted with $-$.).

\subsection{Additional Evaluation}
\subsubsection{Different Architectures}

\begin{table}[t]
\centering
\caption{Performance evaluation on CIFAR-10 dataset. The backbone networks are \textbf{WideResNet-34-10}. Best and second best results are highlighted in boldface and underline, respectively.}
\begin{adjustbox}{max width=\textwidth}
\begin{tabular}{@{}lccc@{}}
\toprule
                                 Method & Standard accuracy  & PGD$_{50}$ & AA \\ \midrule
                                 FBF \citep{wong2019fast}  & 82.1  & 54.4    & 51.3  \\
                                 GAT \citep{sriramanan2020guided}   & 84.7   & 56.1    & 52.1  \\
                                 NuAT \citep{sriramanan2021towards}   & 85.1   & 54.6    & 53.4  \\
                                 PhaseAT (Ours.)  & 88.8   & 62.3    & 59.2  \\ \bottomrule
\end{tabular}
\end{adjustbox}
\label{sup:exp:architecture}
\end{table}

We conduct additional experiments by scaling the PhaseAT backbone networks to verify the effectiveness of PhaseAT on different architectures. We use WideResNet-34-10 architecture instead of PreActResNet-18 to evaluate each baseline on CIFAR-10.The comparison results are listed in Table \ref{sup:exp:architecture}. Similar to the main experiment, we see that PhaseAT achieves the best results amonst all non-iterative methods, demonstrating that PhaseAT can be well scaled to the larger networks.

\subsubsection{Adversarial Robustness against Black-box Attacks}\label{sect:black}

\begin{table}[t]
\centering
\caption{Performance evaluation on CIFAR-10 dataset against two different black-box attacks.}
\begin{tabular}{@{}lcccc@{}}
\toprule
\multirow{2}{*}{Method} & \multirow{2}{*}{Standard accuracy} & \multicolumn{2}{c}{Transfer-based attack} & \multirow{2}{*}{Score-based attack} \\ \cmidrule(lr){3-4}
                  &                                    & VGG-11                 & ResNet-18            &                                     \\ \midrule
FBF \citep{wong2019fast} & 84.0       & 80.5   & 80.6           & 53.5      \\
GAT \citep{sriramanan2020guided}   & 80.5             & 79.8   & 80.3            & 54.1      \\
NuAT \citep{sriramanan2021towards}   & 81.6             & 79.5   & 80.5            & 56.7      \\
PhaseAT (Ours.)             & 86.2             & 83.8   & 85.0           & 76.5      \\  \bottomrule
\end{tabular}
\label{exp:black_cifar}
\end{table}

As DNN models are often hidden from users in real-world applications, the robustness against black-box attacks is also crucial. Among the different kinds of black-box attacks, we consider transfer-based \citep{liu2016delving,papernot2017practical} and score-based attacks. For transfer-based attacks, we use VGG-11 and ResNet-18 as substitute models and construct the attacks using seven steps of PGD \citep{madry2017towards}. For score-based attacks, we adopt square attack \citep{andriushchenko2020square} with 5,000 query budgets, which is a gradient-free attack and one of the strongest attacks in black-box attacks.

Table \ref{exp:black_cifar} shows the robust accuracy against black-box attacks. Similar to white-box attacks, PhaseAT shows better accuracy against both transfer-based and score-based attacks in comparison to other non-iterative methods. In score-based attacks, the difference in performance between others and PhaseAT is particularly noticeable. This can be explained by the stochastic process of PhaseAT because Qin et al. \citep{qin2021random} demonstrate that randomized defense (e.g., Gaussian noise in the inputs) can robustly prevent the model from score-based attacks. This is why we exclude square attack from the AA attack for a fair comparison with other baselines that do not include the stochastic process. Note that the stochastic property can not be circumvented in the black-box scenario because it is infeasible to design adaptive attacks (i.e., EOT attacks) as in the white-box scenario. Comprehensive results show that PhaseAT could be a robust defense strategy against both white-box and black-box attacks.

\begin{table}[ht]
\centering
\caption{Hyper-parameter setting for all baselines.}
\begin{adjustbox}{max width=\textwidth}
\begin{tabular}{@{}clccc@{}}
\toprule
Method                    & Hyper-parameters       & CIFAR-10 (PreActResNet-18)  & ImageNet-100 (ResNet-18) \\ \midrule
\multirow{5}{*}{FBF}      & perturbation $\epsilon$         & 0.031                 & 0.031                           \\
                          & perturbation step size & 0.039                 & 0.039                           \\
                          & learning rate          & 0.1                   & -                           \\
                          & epoch                  & 30                    & -                           \\
                          & batch size             & 256                   & -                           \\ \midrule
\multirow{5}{*}{GAT}      & perturbation $\epsilon$         & 0.031                 & 0.031                           \\
                          & perturbation step size & 0.031                    & 0.031                           \\
                          & learning rate          & 0.1                  & 0.1                           \\
                          & epoch                  & 100                   & 100                           \\
                          & batch size             & 64                   & 64                           \\ \midrule
\multirow{5}{*}{NuAT}     & perturbation $\epsilon$         & 0.031                 & 0.031                           \\
                          & perturbation step size & 0.031                    & 0.031                           \\
                          & learning rate          & 0.1                  & 0.1                           \\
                          & epoch                  & 100                   & 100                           \\
                          & batch size             & 64                   & 64                           \\ \midrule \midrule
\multirow{6}{*}{PGD}      & perturbation $\epsilon$         & 0.031                 & 0.031                           \\
                          & perturbation step size & 0.039                 & 0.039                           \\
                          & number of iterations   & 7                     & -                           \\
                          & learning rate          & 0.1                   & -                           \\
                          & epoch                  & 30                    & -                           \\
                          & batch size             & 256                  & -                           \\ \midrule
\multirow{6}{*}{TRADES}   & perturbation $\epsilon$         & 0.031                & 0.031                           \\
                          & perturbation step size & 0.007                & -                           \\
                          & beta                   & 6.0                  & -                           \\
                          & learning rate          & 0.1                  & -                           \\
                          & epoch                  & 100                  & -                           \\
                          & batch size             & 128                  & -                           \\ \midrule \midrule
\multirow{7}{*}{PhaseDNN} & perturbation $\epsilon$         & 0.031                & 0.031                           \\
                          & perturbation step size & 0.039                & 0.039                          \\
                          & frequency range        & {[}0, 50000{)}       & {[}0, 50000{)}                           \\
                          & number of heads        & 3                    & 3                           \\
                          & learning rate          & 0.1                  & 0.1                           \\
                          & epoch                  & 30                   & 50                           \\
                          & batch size             & 256                  & 128                           \\ \bottomrule
\end{tabular}
\end{adjustbox}
\label{tab:param}
\end{table}

\subsection{Proofs of Theorems 3.1 and 3.2}

\subsubsection{Preliminaries}
Before proving Theorems 3.1 and 3.2 in this section, we start with a detailed explanation of DNNs, and then introduce the mathematical tools required for proof which can be found in standard references (e.g. \citep{stein,wolff,Schlag,Evans}).

\paragraph{Deep Neural Networks.} A DNN with $K$-hidden layers and general activation functions is a vector-valued function $\mathcal T_{\theta}(x): \mathbb{R}^d \rightarrow \mathbb{R}^{m_{K+1}}$
where $m_k$ denotes the number of nodes in the $k$-th layer.
For $1\leq k \leq K+1$, we set ${\boldsymbol W}^{(k)}\in\mathbb{R}^{m_k \times m_{k-1}}$ and ${\boldsymbol b}^{(k)}\in\mathbb{R}^{m_k}$ as the matrices whose entries consist of the weights and biases called parameters.
The parameter vector $\theta$ is then defined as $$\theta=\big(\textrm{vec}({\boldsymbol W}^{(1)}), \textrm{vec}({\boldsymbol b}^{(1)}),\cdots,\textrm{vec}({\boldsymbol W}^{(K+1)}),\textrm{vec}({\boldsymbol b}^{(K+1)})\big)\in\mathbb{R}^M,$$
where $M=\sum_{k=1}^{K+1}(m_{k-1}+1)m_k$ is the number of the parameters. 
Given $\theta\in\mathbb{R}^M$ and an activation function $\sigma:\mathbb{R}\rightarrow \mathbb{R}$, the DNN output $\mathcal T_\theta^{(K+1)}(x):\mathbb{R}^d \rightarrow \mathbb{R}^{m_{K+1}}$ is expressed in terms of composite functions;
setting $\mathcal T_\theta^{(0)}(x)=x$, $\mathcal T_\theta^{(k)}(x):\mathbb{R}^d \rightarrow \mathbb{R}^{m_k}$ is defined recursively as
\begin{equation*}
(\mathcal T_\theta^{(k)}(x))_i = \sigma \big(({\boldsymbol W}^{k}\mathcal T_\theta^{k-1} + {\boldsymbol b}^{k})_i\big), \,\, 1\leq i \leq m_k,\,\, 1\leq k\leq K.
\end{equation*}
We denote the DNN output $\mathcal T_\theta^{(K+1)}(x)={\boldsymbol W}^{(K+1)} \mathcal T_\theta^{{(K)}} + {\boldsymbol b}^{(K+1)}$ by $\mathcal T_\theta(x)$.

\paragraph{The Basic Properties of Fourier Transforms.}
Let $f\in L^1(\mathbb{R}^d)$. 
The Fourier transform of $f$ is defined by
\begin{equation*}
\widehat{f}(\xi)=\int_{\mathbb{R}^d} e^{-2\pi i x\cdot \xi} f(x) dx.
\end{equation*}
Then clearly
\begin{equation}\label{fou}
    \|\widehat f\|_{L^{\infty}} \leq \|f\|_{L^1}.
\end{equation}
Additionally if $\widehat {f} \in L^1(\mathbb{R}^d)$, 
the Fourier inversion holds:
\begin{equation}\label{inv}
f(x)=\int_{\mathbb{R}^d} e^{2\pi i x\cdot \xi} \widehat{f}(\xi) d\xi.
\end{equation}
If $f,g\in L^1 (\mathbb{R}^d)$, then $f\ast g\in L^1(\mathbb{R}^d)$ and
\begin{equation}\label{con}
\widehat{f\ast g} = \hat f \hat g.
\end{equation}
For an $n$-tuple $\alpha=(\alpha_1, \cdots , \alpha_d)$ of nonnegative integers, we denote
$$D^\alpha = \prod_{j=1}^{d} \frac{\partial^{\alpha_j}}{\partial^{\alpha_j}_{x_j}} \quad \textnormal{and} \quad |\alpha|=\sum_{j=1}^d \alpha_j.$$
Then, if $D^\alpha f \in L^1(\mathbb{R}^d)$ whenever $0\leq|\alpha|\leq s$, 
\begin{equation}\label{dif}
\widehat{D^\alpha f} (\xi) = (2\pi i)^{|\alpha|} \xi^\alpha \widehat f(\xi).
\end{equation}

\paragraph{Sobolev Spaces and Gaussian Weights.}
For $s \in \mathbb{N}$, the Sobolev space $W^{s,\infty}(\mathbb{R}^{d})$ is defined as
$$W^{s,\infty}(\mathbb{R}^{d}) = \{ f \in L^{\infty}(\mathbb{R}^d) : D^{\alpha}f \in L^{\infty}(\mathbb{R}^d) \, \textnormal{for all} \, 0\leq|\alpha| \leq s\}$$ equipped with the norm
\begin{equation*}
\| f \|_{W^{s,\infty}(\mathbb{R}^{d})} = \sum_{|\alpha| \leq s} \| D^{\alpha}f\|_{L^\infty (\mathbb{R}^{d})}.
\end{equation*}
We also introduce a Gaussian weight $G_\varepsilon(x)=\varepsilon^{-d}e^{-\pi \varepsilon^{-2} |x|^2}$ for any $\varepsilon>0$ on which the Fourier transform has an explicit form,
\begin{equation}\label{Ga}
\widehat{G_\varepsilon} (\xi) = e^{-\pi \varepsilon^2 |\xi|^2}.    
\end{equation}
The final observation is that $G_\varepsilon$ is an approximate identity with respect to the limit $\varepsilon\rightarrow0$ as in the following well-known lemma:
\begin{lem}\label{delta}
Let $f\in C(\mathbb{R}^d)\cap L^\infty(\mathbb{R}^d)$. Then
\begin{equation}\label{approx}
\lim_{\varepsilon\rightarrow0}\int_{\mathbb{R}^d}
G_\varepsilon(x-y)f(y)dy=f(x)
\end{equation}
for all $x\in\mathbb{R}^d$.
\end{lem}

\subsubsection{Proof of Theorem 3.1}\label{sec:pfthm}
In what follows we may consider a compact domain $\Omega$ instead of $\mathbb{R}^d$ because the input data $\{x_j\}_{j=0}^{N-1}$ used for training is sampled from a bounded region. 

For a discrete input data $\{x_j\}_{j=0}^{N-1}$, we now recall the total loss in adversarial training from Section 3.1:
\begin{equation}\label{total0}
L(\theta) = \frac{1}{N} \sum_{j=0}^{N-1} \ell(\mathcal{T}_{\theta}\circ\mathcal A,g)(x_j).
\end{equation}
From the continuity of $\mathcal T_\theta$ and $g$ in the compact domain $\Omega$, we note that $\ell(\mathcal T_\theta \circ A,g)$ is continuous and bounded for general loss functions such as mean-squared error loss and cross-entropy loss.
Then we can apply Lemma \ref{delta} to deduce
\begin{align}\label{total}
\nonumber
L(\theta)&=\lim_{\varepsilon \rightarrow 0}\frac{1}{N} \sum_{j=0}^{N-1}\int_{\mathbb{R}^d} G_{\varepsilon}(x_j-x) \ell(\mathcal{T}_{\theta}\circ\mathcal A,g)(x)dx\\
&=\lim_{\varepsilon \rightarrow 0}\frac{1}{N} \sum_{j=0}^{N-1} \big(G_{\varepsilon} \ast\ell(\mathcal{T}_{\theta}\circ\mathcal A,g)\big)(x_j).
\end{align}
Using the properties $G_{\varepsilon} \in L^1(\mathbb{R}^d)$ and $\ell(\mathcal T_{\theta} \circ \mathcal A,g) \in L^1(\mathbb{R}^d)$, we then derive from Eq.~\ref{con} and Eq.~\ref{Ga} that 
\begin{equation*}
G_{\varepsilon}\ast\ell(\mathcal{T}_{\theta}\circ\mathcal A,g) \in L^1(\mathbb{R}^d)
\end{equation*}
and
\begin{equation}\label{df}
\reallywidehat{G_{\varepsilon}\ast\ell(\mathcal{T}_{\theta}\circ\mathcal A,g)} (\xi) =e^{-\pi \varepsilon^2 |\xi|^2}\, \reallywidehat{\ell(\mathcal{T}_{\theta}\circ\mathcal A,g)}(\xi).
\end{equation}
Note here that by Eq.~\ref{fou}
\begin{equation*}
\|e^{-\pi \varepsilon^2 |\xi|^2} \, \reallywidehat{\ell(\mathcal{T}_{\theta}\circ\mathcal A,g)}(\xi)\|_{L^1} \leq \|\reallywidehat{\ell(\mathcal{T}_{\theta}\circ\mathcal A,g)}\|_{L^{\infty}} \|e^{-\pi \varepsilon^2 |\xi|^2}\|_{L^1} \leq C \|\ell(\mathcal{T}_{\theta}\circ\mathcal A,g)\|_{L^1} < \infty. 
\end{equation*}
Hence the Fourier inversion Eq.~\ref{inv} together with Eq.~\ref{df} implies
\begin{equation}\label{conv}
\big(G_{\varepsilon}\ast\ell(\mathcal{T}_{\theta}\circ\mathcal A,g)\big)(x_j)
=\int_{\mathbb{R}^d}e^{2\pi i x_j \cdot \xi}e^{-\pi \varepsilon^2 |\xi|^2}\, \reallywidehat{\ell(\mathcal{T}_{\theta}\circ\mathcal A,g)}(\xi) d\xi.
\end{equation}
Substituting Eq.~\ref{conv} into the right-hand side of Eq.~\ref{total}, 
we immediately obtain 
\begin{equation}\label{arg}
L(\theta)=\lim_{\varepsilon \rightarrow 0} \frac{1}{N} \sum_{j=0}^{N-1}\int_{\mathbb{R}^d} e^{2\pi i x_j \cdot\xi }\, e^{-\pi \varepsilon^2 |\xi|^2} \reallywidehat{\ell\big(\mathcal T_{\theta}\circ \mathcal{A},g\big)}(\xi) d\xi,
\end{equation}
as desired. This completes the proof.

\subsubsection{Proof of Theorem 3.2}
\paragraph{Representing $\nabla_\theta L(\theta)$ in the frequency domain.}
To begin with, we represent $\nabla_\theta L(\theta)$ in the frequency domain in the same way as in Section \ref{sec:pfthm}.
By differentiating both sides of Eq.~\ref{total0} with respect to $\theta$ and using Lemma \ref{delta}, we first see  
\begin{align}\label{gra}
\nonumber
\nabla_\theta L(\theta) &= \frac{1}{N} \sum_{j=0}^{N-1} \nabla_\theta\ell(\mathcal{T}_{\theta}\circ\mathcal A,g)(x_j)\\
&=\lim_{\varepsilon\rightarrow 0}\frac{1}{N} \sum_{j=0}^{N-1}\int_{\mathbb{R}^d} G_\varepsilon (x_j-x)\nabla_{\theta}\ell(\mathcal{T}_{\theta}\circ\mathcal A,g)(x) dx
\end{align}
if $\nabla_\theta \ell(\mathcal T_\theta \circ \mathcal A,g)$ is continuous and bounded.
Since $\ell(\mathcal T_\theta \circ \mathcal A,g)$ is differentiable with respect to the first argument (as mentioned in Section 3.1) and $\mathcal T_\theta$ is differentiable with respect to $\theta$ for general activation functions such as ReLU, eLU, tanh and sigmoid, the continuity is generally permissible, and thus the boundedness follows also from compact domain.
In fact, the ReLU activation function is not differentiable at the origin and neither is $\mathcal T_\theta$ on a certain union of hyperplanes;
for example, when considering $1$-hidden layer neural network with $m_1$ nodes and $1$-dimensional output, the output is 
$$\mathcal T_\theta (x) = \sum_{i=1}^{m_1} w_i^{(2)} \sigma (\boldsymbol W_i^{(1)} \cdot x + \boldsymbol b_i^{(1)}),\quad w_i^{(2)},b_i^{(1)} \in \mathbb{R}, \,\boldsymbol W_i^{(1)}\in \mathbb{R}^d$$
%$(\mathcal T_\theta(x))_i = w \sigma\big((\boldsymbol W^{(1)} \cdot x + \boldsymbol b^{(1)})_i \big)+b$ 
and the set of non-differentiable points is a union of hyperplanes given by $\{x\in \mathbb{R}^d:\boldsymbol W_i^{(1)} \cdot x + \boldsymbol b_i^{(1)}=0,\, 1\leq i \leq m_1\}$. But the $d$-dimensional volume of such thin sets is zero and thus they may be excluded from the integration region in Eq.~\ref{approx} when applying Lemma \ref{delta} to obtain Eq.~\ref{gra} for the case of ReLU.

Just by replacing $L(\theta)$ with $\nabla_\theta L(\theta)$ in the argument employed for the proof of Eq.~\ref{arg} and repeating the same argument, it follows now that
\begin{equation}\label{loss} 
\nabla_\theta L(\theta)=\lim_{\varepsilon \rightarrow 0} \frac{1}{N} \sum_{j=0}^{N-1}\int_{\mathbb{R}^d} e^{2\pi i x_j \cdot\xi }\, e^{-\pi \varepsilon^2 |\xi|^2} \reallywidehat{\nabla_\theta \ell\big(\mathcal T_{\theta}\circ \mathcal{A},g\big)}(\xi) d\xi.
\end{equation}
We then pull $\nabla_\theta$ to the outside of the integration in Eq.~\ref{loss}, and recall $L_{\leq\eta}(\theta)$ and $L_{\geq\eta}(\theta)$ 
from Section 3.1, contributed by low and high frequencies in the loss, to see
\begin{equation}\label{app}
\nabla_\theta L(\theta) \approx \nabla_\theta L_{\leq\eta}(\theta) + \nabla_\theta L_{\geq\eta}(\theta).
\end{equation}
This approximation is more and more accurate as $\varepsilon$ diminishes smaller in Eq.~\ref{loss}, and the size of $\varepsilon$ will be later determined inversely proportional to the number of dataset $N$ or dimension $d$ to consider a natural approximation reflecting the discrete experimental setting.

\paragraph{Estimating $\nabla_\theta L_{\geq\eta}(\theta)$ in terms of $\eta$.}
Now we show that for the $i$-th element of $\nabla_\theta L_{\geq\eta}(\theta)$
\begin{equation}\label{el}
\Big|\frac{\partial L_{\geq\eta}(\theta)}{\partial\theta_i}\Big| \leq C\max(N, d^d)\,\eta^{-2s}
\end{equation}
which implies
\begin{equation*}
|\nabla_{\theta}L_{\geq\eta}(\theta)|= \Big(\sum_{\theta} \Big|\frac{\partial L_{\geq\eta}(\theta)}{\partial\theta_i}\Big|^2\Big)^{1/2} \leq C \max(N, d^d)\, \eta^{-2s}.
\end{equation*}
By Eq.~\ref{app} and this bound, we get 
\begin{equation*}
|\nabla_\theta L(\theta) - \nabla_\theta L_{\leq\eta}(\theta)|\approx  |\nabla_\theta L_{\geq\eta}(\theta)|\leq C \max(N, d^d)\, \eta^{-2s}
\end{equation*}
which completes the proof of Theorem 3.2.

To show Eq.~\ref{el}, we first use the chain rule to calculate 
\begin{align*}
\frac{\partial L_{\geq\eta}(\theta)}{\partial \theta_i}= \frac{1}{N} \sum_{j=0}^{N-1} \int_{|\xi|\geq \eta} e^{2\pi i x_j \cdot \xi} e^{-\pi \varepsilon^2 |\xi|^2}\reallywidehat{\nabla_{\mathcal T_\theta}\ell(\mathcal T_{\theta}\circ \mathcal A,g)\cdot \frac{\partial(\mathcal T_\theta\circ \mathcal A)}{\partial\theta_i}}(\xi) d\xi.
\end{align*}
Since $\eta \leq \langle\xi\rangle:=\sqrt{1+|\xi|^2}$ for all $0<\eta \leq |\xi|$, we then see that for $s\in\mathbb{N}$
\begin{align}\label{1}
\nonumber
\bigg|\frac{\partial L_{\geq\eta}(\theta)}{\partial \theta_i}\bigg| &\leq \frac{1}{N} \sum_{j=0}^{N-1} \eta^{-2s}\int_{|\xi|\geq \eta} e^{-\pi \varepsilon^2 |\xi|^2}\Big|\langle\xi\rangle^{2s}\reallywidehat{\nabla_{\mathcal T_\theta}\ell(\mathcal T_{\theta}\circ \mathcal A,g)\cdot \frac{\partial(\mathcal T_\theta\circ \mathcal A)}{\partial\theta_i}}(\xi)\Big|d\xi\\
&\leq \frac{1}{N} \sum_{j=0}^{N-1} \eta^{-2s}\Big\|\langle\xi\rangle^{2s}\reallywidehat{\nabla_{\mathcal T_\theta}\ell(\mathcal T_{\theta}\circ \mathcal A,g)\cdot \frac{\partial(\mathcal T_\theta\circ \mathcal A)}{\partial\theta_i}}(\xi)\Big\|_{L^{\infty}} \|e^{-\pi \varepsilon^2 |\xi|^2}\|_{L^1}.
\end{align}
By a change of variables $\varepsilon \xi \rightarrow \xi$, we note
\begin{equation*}
\|e^{-\pi \varepsilon^2|\xi|^2}\|_{L^1} = \varepsilon^{-d}\int_{\mathbb{R}^d} e^{-\pi |\xi|^2} d\xi \leq C \varepsilon^{-d}.
\end{equation*}
Hence, if we show that the $L^\infty$-norm in Eq.~\ref{1} is finite, then
\begin{equation*}
\bigg|\frac{\partial L_{\geq\eta}(\theta)}{\partial \theta_i}\bigg|\leq C \eta^{-2s}\varepsilon^{-d}.
\end{equation*}
Finally, if we take $\varepsilon= \min\{1/\sqrt[d]{N}, 1/ d\}$ for large $N, d$, we conclude  
\begin{equation}\label{still}
\bigg|\frac{\partial L_{\geq\eta}(\theta)}{\partial \theta_i}\bigg|\leq C \max\{N, d^d\}\eta^{-2s}
\end{equation}
as desired.

Now all we have to do is to bound the $L^\infty$-norm in Eq.~\ref{1}. 
Using the simple inequalities
\[\langle\xi\rangle\leq 1+|\xi|,\quad(1+|\xi|)^M \leq C \sum_{|\alpha|\leq M} |\xi^\alpha|,\] and Eq.~\ref{dif}, Eq.~\ref{fou} in turn, we first see 
\begin{align*}
\Big\|\langle\xi\rangle^{2s}\reallywidehat{\nabla_{\mathcal T_{\theta}}\ell(\mathcal T_{\theta}\circ \mathcal A,g)\cdot \frac{\partial(\mathcal T_\theta\circ \mathcal A)}{\partial\theta_i}}\Big\|_{L^{\infty}} &\leq \Big\|(1+|\xi|)^{2s}\reallywidehat{\nabla_{\mathcal T_{\theta}}\ell(\mathcal T_{\theta}\circ \mathcal A,g)\cdot \frac{\partial(\mathcal T_\theta\circ \mathcal A)}{\partial\theta_i}}\Big\|_{L^{\infty}}\\
\nonumber
&\leq C\sum_{|\alpha|\leq 2s}\Big\|\xi^{\alpha}\reallywidehat{\nabla_{\mathcal T_{\theta}}\ell(\mathcal T_{\theta}\circ \mathcal A,g)\cdot \frac{\partial(\mathcal T_\theta\circ \mathcal A)}{\partial\theta_i}}\Big\|_{L^{\infty}}\\
\nonumber
&\leq C\sum_{|\alpha|\leq 2s}\Big\|\reallywidehat{D^{\alpha}\big(\nabla_{\mathcal T_{\theta}}\ell(\mathcal T_{\theta}\circ \mathcal A,g)\cdot \frac{\partial(\mathcal T_\theta\circ \mathcal A)}{\partial\theta_i}\big)}\Big\|_{L^{\infty}}\\
&\leq C\sum_{|\alpha|\leq 2s}\Big\|D^{\alpha}\big(\nabla_{\mathcal T_{\theta}}\ell(\mathcal T_{\theta}\circ \mathcal A,g)\cdot \frac{\partial(\mathcal T_\theta\circ \mathcal A)}{\partial\theta_i}\big)\Big\|_{L^1}.
\end{align*}
By Leibniz's rule we then bound the $L^1$-norm in the above as 
\begin{equation*}
\Big\|D^{\alpha}\big(\nabla_{\mathcal T_{\theta}}\ell(\mathcal T_{\theta}\circ \mathcal A,g)\cdot \frac{\partial(\mathcal T_\theta\circ \mathcal A)}{\partial\theta_i}\big)\Big\|_{L^1} \leq C \sum_{|\alpha_1|+|\alpha_2|\atop=|\alpha|}\Big\|D^{\alpha_1}\nabla_{\mathcal T_{\theta}}\ell(\mathcal T_{\theta}\circ \mathcal A,g)\cdot D^{\alpha_2}\frac{\partial(\mathcal T_\theta\circ \mathcal A)}{\partial\theta_i} \Big\|_{L^1}.    
\end{equation*}

When $|\alpha|\leq s-1$, $|\alpha_1|\leq s$ and $|\alpha_2|+1\leq s$, 
and then the $L^1$-norm in the right-hand side is generally finite since $\sigma\in W^{s,\infty}(\mathbb{R})$, $g \in W^{s,\infty}(\mathbb{R}^d)$, and the $L^1$-norm may be taken over the compact domain $\Omega$;
for example, $\ell(\mathcal T_\theta \circ \mathcal A,g)(x)=|(\mathcal T_\theta \circ \mathcal A)(x)-g(x)|^2$ and 
\begin{equation*}
\nabla_{\mathcal T_\theta} \ell(\mathcal T_\theta \circ \mathcal A,g)(x)=2((\mathcal T_\theta \circ \mathcal A)(x)-g(x))\end{equation*}
for mean-squared error loss.
Since $(\mathcal T_\theta \circ \mathcal A)(x)$ is expressed as compositions of $\sigma$, the regularity of $\nabla_{\mathcal T_\theta} \ell (\mathcal T_\theta \circ \mathcal A, g)(x)$ is exactly determined by that of $\sigma$ and $g$.
Namely, 
\begin{equation}\label{theat}
\nabla_{\mathcal T_\theta} \ell (\mathcal T_\theta \circ \mathcal A, g)(x)\in W^{s,\infty}(\mathbb{R}^d), \quad  (\mathcal T_\theta \circ \mathcal A)(x) \in W^{s,\infty}(\mathbb{R}^d),
\end{equation}
from which the $L^1$-norm taken over the compact domain $\Omega$ is finite
since $|\alpha_1|\leq s$ and $|\alpha_2|+1\leq s$.

On the other hand, when $s\leq|\alpha|\leq 2s$ we set $|\alpha|=s+j$ with $0\leq j \leq s$.
Firstly, if $0\leq|\alpha_1|\leq s$ (and so $j\leq|\alpha_2|\leq s+j$ since $|\alpha|=|\alpha_1|+|\alpha_2|$), then we bound
\begin{align*}
\Big\|D^{\alpha_1} \nabla_{\mathcal T_{\theta}}\ell(\mathcal T_{\theta}\circ \mathcal A,g)\cdot& D^{\alpha_2} \frac{\partial(\mathcal T_{\theta}\circ \mathcal A)}{\partial\theta_i}\Big\|_{L^1}\leq \big\|D^{\alpha_1} \nabla_{\mathcal T_{\theta}}\ell(\mathcal T_{\theta}\circ \mathcal A,g)\big\|_{L^\infty} \Big\|D^{\alpha_2} \frac{\partial(\mathcal T_{\theta}\circ \mathcal A)}{\partial\theta_i}\Big\|_{L^1}.
\end{align*}
Here, by Eq.~\ref{theat}, the $L^\infty$-norm in the right-hand side is finite since $|\alpha_1|\leq s$, while the finiteness of $L^1$-norm follows from the fact that $D^{\beta}\sigma \in L^1(\mathbb{R})$ where $|\beta|=|\alpha_2|+1$. This fact is indeed valid for general activation functions such as ReLU, eLU, tanh and sigmoid since the $L^1$-norm may be taken over the compact domain $\Omega$.
Finally, if $s+1\leq|\alpha_1|\leq s+j$ (and so $0\leq|\alpha_2|\leq j-1$), then we bound this time
\begin{align*}
\Big\|D^{\alpha_1} \nabla_{\mathcal T_{\theta}}\ell(\mathcal T_{\theta}\circ \mathcal A,g)\cdot& D^{\alpha_2} \frac{\partial(\mathcal T_{\theta}\circ \mathcal A)}{\partial\theta_i}\Big\|_{L^1}\leq \big\|D^{\alpha_1} \nabla_{\mathcal T_{\theta}}\ell(\mathcal T_{\theta}\circ \mathcal A,g)\big\|_{L^1} \Big\|D^{\alpha_2} \frac{\partial(\mathcal T_{\theta}\circ \mathcal A)}{\partial\theta_i}\Big\|_{L^\infty}.
\end{align*}
Here, the $L^\infty$-norm in the right-hand side is finite by Eq.~\ref{theat} since $|\alpha_2|+1\leq j\leq s$.
The finiteness of $L^1$-norm also comes from $D^{\beta}\sigma\in L^1(\mathbb{R})$ and $D^{\beta}g \in L^1(\mathbb{R}^d)$ with $s+1\leq|\beta|\leq 2s$.
Here, the condition $D^{\beta}\sigma\in L^1(\mathbb{R})$ is valid generally as above, and the bound Eq.~\ref{still} is still valid with $\eta^{-s}$ even if the condition $D^{\beta}g \in L^1(\mathbb{R}^d)$ is not required. 
This is the case $j=0$ in the proof.

\end{document}